\documentclass[sn-basic]{sn-jnl}
\usepackage[utf8]{inputenc}
\usepackage{graphicx}
\usepackage{rotating,pdflscape}
\usepackage[above,below]{placeins}
\usepackage[english]{babel}
\usepackage{xspace}
\usepackage{csquotes}
\usepackage{amssymb}
\usepackage{calc}
\usepackage{enumitem}
\setlist[itemize]{align=left, labelwidth=\widthof{$\bullet$}, labelsep=0.5em, itemindent=0pt, labelindent=0pt, leftmargin=!, topsep=1ex}
\setlist[enumerate]{align=left, labelwidth=\widthof{3.}, labelsep=0.5em, itemindent=0pt, labelindent=0pt, leftmargin=!, topsep=1ex}
\setlist[description,1]{topsep=1ex} 
\newlist{subparagraphs}{description}{1}
\setlist[subparagraphs]{topsep=1ex}

\usepackage{hyperref,xcolor}
\hypersetup{
    pdfauthor={G. Schwalbe and B. Finzel},
    pdftitle={A Taxonomy for XAI},
}
\usepackage{tabularx,tabulary,booktabs}
\usepackage{ltablex}

\addto\extrasenglish{%
}

\makeatletter
\newcommand*{\appenddot}[1]{\@ifnextchar{.}{#1}{}#1.}
\newcommand*{\etc}{\@abbrdot{etc}\@\xspace}
\newcommand*{\forexample}{\textit{\appenddot{e.g}}\@\xspace}
\newcommand*{\idest}{\textit{\appenddot{i.e}}\@\xspace}
\newcommand*{\etal}{et \appenddot{al}\@\xspace}
\newcommand*{\cf}{\appenddot{cf}\@\xspace}
\newcommand{\abbr}[2]{\emph{#1}=#2}

\definecolor{OurGray}{rgb}{0.35,0.35,0.35}
\newcommand*{\parhead}[1]{\textbf{\appenddot{#1}}\@\xspace}
\newenvironment{examples}{\strut\par\color{OurGray}%
    \begin{quote}
    \textit{\appenddot{Examples}\@\xspace}\nopagebreak%
    \newcommand{\examplemethod}[1]{%
      \par\noindent{\normalfont\color{black}\parhead{##1}}
    }%
    }%
    {\end{quote}}
\newcommand*{\inlineexamplemethod}[1]{\textbf{\color{black}#1}}

\makeatletter
\newcommand*{\DivideLengths}[2]{%
  \strip@pt\dimexpr\number\numexpr\number\dimexpr#1\relax*65536/\number\dimexpr#2\relax\relax sp\relax
}
\makeatother
\edef\thetaxonomygraphicscale{\DivideLengths{.95\textwidth}{\pdfpagewidth}}

\newcommand{\rot}[1]{\rotatebox{90}{\parbox{10ex}{\textbf{#1}}}}

\newcommand{\secseprule}[1]{\cmidrule[\heavyrulewidth]{1-#1}} 

\let\longcite\cite
\let\cite\citep

\jyear{2022}%
\title[A Taxonomy for XAI]{A Comprehensive Taxonomy for Explainable Artificial Intelligence: A Systematic Survey of Surveys on Methods and Concepts}
\author*[1,2]{\fnm{Gesina} \sur{Schwalbe}}\email{gesina.schwalbe@continental-corporation.com}
\equalcont{These authors contributed equally to this work.}
\author[2]{\fnm{Bettina} \sur{Finzel}}\email{bettina.finzel@uni-bamberg.de}
\equalcont{These authors contributed equally to this work.}
\affil*[1]{\orgname{Continental AG}, \orgaddress{\city{Regensburg}, \country{Germany}}}
\affil[2]{\orgdiv{Cognitive Systems Group}, \orgname{University of Bamberg}, \orgaddress{\city{Bamberg}, \country{Germany}}}
\abstract{
In the meantime, a wide variety of terminologies, motivations, approaches,
and evaluation criteria have been developed within the research field of
explainable artificial intelligence (XAI).
With the amount of XAI methods vastly growing, a taxonomy of methods is needed
by researchers as well as practitioners:
To grasp the breadth of the topic, compare methods, and to select the right XAI method
based on traits required by a specific use-case context.
Many taxonomies for XAI methods of varying level of detail and depth can be
found in the literature. While they often have a different focus, they also exhibit
many points of overlap.
This paper unifies these efforts and provides a complete taxonomy of XAI methods with respect to notions present in the current state of research.
In a structured literature analysis and meta-study, we identified and reviewed
more than 50 of the most cited and current surveys on XAI methods, metrics, and method traits.
After summarizing them in a survey of surveys,
we merge terminologies and concepts of the articles into a unified structured taxonomy.
Single concepts therein are illustrated by more than 50 diverse selected example methods in total,
which we categorize accordingly.
The taxonomy may serve both beginners, researchers, and practitioners as a
reference and wide-ranging overview of XAI method traits and aspects.
Hence, it provides foundations for targeted, use-case-oriented, and context-sensitive future research.
}


\begin{document}
\maketitle
\bmhead{Acknowledgments}
We would like to thank Christian Hellert as well as the anonymous reviewers 
for their detailed and valuable feedback.
The research leading to these results is partly funded by the BMBF ML-3 project
Transparent Medical Expert Companion (TraMeExCo), FKZ 01IS18056 B, 2018–2021,
and by the German Federal Ministry for Economic Affairs and Energy
within the project \enquote{KI Wissen – Automotive AI powered by Knowledge}.
We would like to thank the consortium for the successful cooperation.


\section{Introduction}

Machine learning (ML) models offer the great benefit that they can deal with
hardly specifiable problems as long as these can be exemplified by data samples.
This has opened up a lot of opportunities for promising automation and assistance
systems, like highly automated driving, medical assistance systems, text summaries
and question-answer systems, just to name a few.
However, many types of models that are automatically learned from data will not
only exhibit high performance, but also be black-box---\idest, they hide information on the
learning progress, internal representation, and final processing in a format
not or hardly interpretable by humans.

There are now diverse use-case specific motivations for allowing humans to
\emph{understand} a given software component, \idest, to build up a mental
model approximating the algorithm in a certain way.
This starts with legal reasons, like the General Data Protection Regulation~\cite{goodman_european_2017}
adopted by the European Union in recent years.
Another example are domain specific standards, like the functional safety standard
ISO\,26262~\cite{iso/tc22/sc32_iso_2018a} requiring accessibility of software components
in safety-critical systems. This is even detailed to an explicit requirement for explainability
of machine learning based components in the ISO/TR\,4804~\cite{iso/tc22roadvehicles_iso_2020}
draft standard.
Many further reasons of public interest, like fairness or security, as well as
business interests like ease of debugging, knowledge retrieval, or appropriate user trust
have been identified \cite{arrieta_explainable_2020,langer_what_2021}.
This need to translate behavioral or internal aspects of black-box algorithms
into a human interpretable form gives rise to the broad research field of
explainable artificial intelligence (XAI).

In recent years, the topic of XAI methods has received an exponential boost in
research interest \cite{arrieta_explainable_2020,linardatos_explainable_2021,adadi_peeking_2018,zhou_evaluating_2021}.
For practical application of XAI in human-AI interaction systems, it is important to 
ensure a choice of XAI method(s) appropriate for the corresponding use case.
Without question, thorough use-case analysis including the main goal and derived requirements
is one essential ingredient here \cite{langer_what_2021}.
Nevertheless, we argue that a necessary foundation for choosing correct requirements is
a complete knowledge of the different
\emph{aspects} (traits, properties) of XAI methods that may influence their applicability.
Well-known aspects are, \forexample,
portability, \idest, whether the method requires access to the model internals or not, 
or locality, \idest, whether single predictions of a model are explained or some global properties.
As will become clear from our literature analysis in \autoref{sec:taxonomy},
this only just scratches the surface of aspects of XAI methods that are relevant for practical application.

This paper aims to
(1)~help beginners in gaining a good initial overview and starting point for a deeper dive,
(2)~support practitioners seeking a categorization scheme for choosing an appropriate XAI method for their use-case, and
(3)~to assist researchers in identifying desired combination of aspects that have not or little been considered so far.
For this, we provide a complete collection and a structured overview in the form of a
taxonomy of XAI method aspects in \autoref{sec:taxonomy}, together with method examples for each aspect.
The method aspects are obtained from an extensive literature survey on categorization
schemes for explainability and interpretability methods,
resulting in a meta-study on XAI surveys presented in \autoref{sec:sos}.
Other than similar work, we do not aim to provide a survey on XAI methods.
Hence, our taxonomy is not constructed as a means of a sufficient chapter scheme. Instead, we try to compile a taxonomy that is complete with respect to existing valuable work.
We believe that this gives a good starting point for an in-depth understanding
of sub-topics of XAI research, and research on XAI methods themselves.

Our main contributions are:
\begin{itemize}
    \item \emph{Complete XAI method taxonomy:}
        A structured, detailed, and deep taxonomy of XAI method aspects (see \autoref{fig:taxonomy});
        In particular, the taxonomy is complete with respect to application relevant
        XAI method and evaluation aspects that have so far been considered in literature, according to our structured literature search.
    \item \emph{Extensive XAI method meta-study:}
        A survey-of-surveys of more than 50 works on XAI related topics that may serve
        to pick the right starting point for a deeper dive into XAI and XAI sub-fields.
        To our knowledge, this is the most extensive and detailed meta-study specifically on XAI and XAI methods
        available so far.
    \item \emph{Broad XAI method review:}
        A large collection, review, and categorization of more than 50 hand-picked diverse XAI methods
        (see \autoref{tab:methods.overview}).
        This evidences the practical applicability of our taxonomy structure and the identified method aspects.
\end{itemize}

The rest of the paper reads as follows:
In \autoref{sec:background} we reference some related work, recapitulate major milestones in the history of XAI and provide important definitions of terms.
This is meant for those readers less familiar with the topic of XAI.
After that, we detail our systematic review approach in \autoref{sec:approach}.
The results of the review are split into the following chapters:
The detailed review of the selected surveys is presented in \autoref{sec:sos},
including their value for different audiences and research focus;
and \autoref{sec:taxonomy} details collected XAI method aspects and our proposal of a taxonomy thereof.
Each considered aspect is accompanied by illustrative example methods, a summary of which can be found in \autoref{tab:methods.overview}. We conclude our work in \autoref{sec:conclusion}.


\section{Background}
\label{sec:background}

This chapter gives some background regarding
related work (\autoref{sec:relatedwork}), and
the history of XAI including its main milestones (\autoref{sec:history}).
Lastly, \autoref{sec:def} introduces some basic terms and definitions used throughout this work.
Experienced readers may skip the respective subsections.

\subsection{Related work}  
\label{sec:relatedwork}

\paragraph*{Meta-studies on XAI methods}
Short meta-studies collecting surveys of XAI methods are often contained in the
related work section of XAI reviews like \cite[Sec.\,2.3]{linardatos_explainable_2021}.
These are, however, by nature restricted in length and usually concentrate on
most relevant and cited reference works like \longcite{gilpin_explaining_2018,adadi_peeking_2018,arrieta_explainable_2020}.
We here instead consider a very broad and extensive collection of surveys.
%
By now, also few dedicated meta-studies can be found in literature.
One is the comprehensive and recent systematic literature survey
conducted by \longcite[Chap.~4]{vilone_explainable_2020}.
This reviews 53 survey articles related to XAI topics,
which are classified differentially according to their focus.
Their literature analysis reaches further back in time and targets even more general topics in XAI than ours.
Hence, several of their included surveys are very specific, quite short, and do not provide any structured
notions of taxonomies. Also, the focus of their review lies in research topics,
while the slightly more detailed survey-of-surveys presented here also takes into account
aspects that relate to suitability for beginners and practitioners.
%
Another dedicated survey-of-surveys is the one by \longcite{chatzimparmpas_survey_2020}.
With 18 surveys on visual XAI this is smaller compared to ours, and has quite a different focus:
They review the intersection of XAI with visual analytics.
Lastly, due to their publication date they only include surveys from 2018 or earlier,
which misses on important recent work as will be seen in \autoref{fig:surveydist}.
%
The same holds for the very detailed DARPA report by \longcite{mueller_explanation_2019} from 2019. Hence, both papers miss many recent works covered in our meta-study (\cf~\autoref{fig:surveydist}).
Further, the DARPA report has a very broad focus, also including sibbling research fields related to XAI. And, similar to Vilone et al., they target researchers,
hence give no estimation of what work is especially suitable for beginners or practitioners.

\paragraph*{(Meta-)Studies on XAI method traits}
Related work on taxonomies is mostly shallow, focused on a sub-field, or is
purely functional in the sense that taxonomies merely serve as a survey chapter structure.
A quite detailed, but less structured collection of XAI method traits can be found
in the courses of discussion in \longcite{murdoch_definitions_2019,carvalho_machine_2019,burkart_survey_2021,mueller_explanation_2019,li_quantitative_2020}.
The focus of each will be discussed later in \autoref{sec:sos}.
But despite their detail, we found that each article features unique aspects that are not included in the others.
The only systematic meta-review on XAI method traits known to us is the mentioned
survey by \longcite{vilone_explainable_2020}. They, however, generally review
notions related to XAI, not concretely XAI method traits. Only a shallow taxonomy is
derived from parts of the retrieved notions (\cf~\cite[Sec.\,5]{vilone_explainable_2020}). 

\paragraph*{Use-case analysis}
A related, wide and important field which is out-of-scope of this paper is that of use-case
and requirements analysis. This, \forexample, includes analysis of the background of the
explanation receiver \cite{miller_explanation_2019}.
Instead, we here concentrate on finding aspects of the XAI methods themselves
that may be used for requirements analysis and formulation,
\forexample, whether the explainability method must be model-agnostic or the
amount of information conveyed to the receiver must be small.
For further detail on use-case specific aspects the reader may be referred to one of
the following surveys further discussed in \autoref{sec:sos}
\cite{miller_explanation_2019,guidotti_survey_2018,gleicher_framework_2016,langer_what_2021,arrieta_explainable_2020}.


\subsection{History of XAI}
\label{sec:history}

XAI is not a new topic, although the number of papers published in recent years might suggest it. The abbreviation XAI for the term explainable artificial intelligence was first used by \longcite{vanlent_explainable_2004} (see~\cite[Sec.\,2.3]{carvalho_machine_2019}). According to \longcite{belle2017logic}, the first mention of the underlying concept already dates back to the year 1958, when McCarthy described his idea of how to realize AI systems. In his seminal paper he promoted a declarative approach, based on a formal language and a problem-solving algorithm that operates on data represented in the given formal language. Such an approach could be understood by humans and the system's behaviour could be adapted, if necessary \cite{mccarthy1958}. Basically, McCarthy described the idea of inherently transparent systems that would be explainable by design.

Inherently transparent approaches paved the way for expert systems that were designed to support human decision-makers (see~\cite{jackson1998intro}). Due to the big challenge to integrate, often implicit, expert knowledge, these systems lost their popularity in the 90's of the last century. Meanwhile, deep neural networks (DNNs) have become a new and powerful approach to solve sub-symbolic problems. The first approaches aiming at making neural network decisions transparent for debugging purposes date back to the mid 90's. For example in 1992, Craven and Shavlik presented several methods to visualize numerical data, such as decision surfaces \cite{craven1992visualizing}.

Due to the introduction of the new European General Data Protection Regulation (GDPR) in May 2018, transparency of complex and opaque approaches, such as neural networks, took on a new meaning. Researchers and companies started to develop new AI frameworks, putting more emphasis on the aspect of accountability and the \enquote{right of explanation} \cite{goodman_european_2017,council2017statement}. Besides debugging and making decisions transparent to developers or end-users, decisions of a system now also had to be comprehensible for further stakeholders.
According to \longcite{adadi_peeking_2018} the term XAI gained popularity in the research community after the Defense Advanced Research Projects Agency (DARPA) published its paper about explainable artificial intelligence, see~\longcite{gunning2019darpa}.

In their definition, XAI efforts aim for two main goals. The first one is to create machine learning techniques that produce models that can be explained (their decision-making process as well as the output), while maintaining a high level of learning performance. The second goal is to convey a user-centric approach, in order to enable humans to understand their artificial counterparts. As a consequence, XAI aims for increasing the trust in learned models and to allow for an efficient partnership between human and artificial agents \cite{gunning2019darpa}.
In order to reach the first goal, DARPA proposes three strategies: deep explanation, interpretable models and model induction, which are defined in \autoref{sec:def}. Among the most prominent XAI methods that implement this goal for deep learning, especially in computer vision, are for example LIME \cite{ribeiro_why_2016}, LRP \cite{bach_pixelwise_2015} and Grad-CAM \cite{selvaraju_gradcam_2017}.
The second, more user-centric, goal defined by DARPA requires a highly inter-disciplinary perspective. This is based on fields such as computer science, social sciences as well as psychology in order to produce more explainable models, suitable explanation interfaces, and to communicate explanations effectively under consideration of psychological aspects.

\longcite{miller_explanation_2019}, \forexample, made an important contribution to the implementation of the user-centric perspective with his paper on artificial intelligence from the viewpoint of the social sciences. He considers philosophical, psychological, and interaction-relevant aspects of explanation, examining different frameworks and requirements. An important turn toward a user-centric focus of XAI was also supported by \longcite{rudin_stop_2019} in her paper from 2019, where she argues for using inherently interpretable models instead of opaque models, motivated by the right to explanation and the societal impact of intransparent models.

Another milestone in the development of XAI is the turn toward evaluation metrics for explanations \cite{mueller_principles_2021}. The XAI community now acknowledges more in depth that it is not enough to generate explanations, but that it is also crucial to evaluate how good they are with respect to some formalized measure.

\subsection{Basic definitions}
\label{sec:def}

This section introduces some basic terms related to XAI which are used throughout this paper.
Detailed definitions of the identified XAI method aspects will be given in \autoref{sec:taxonomy}.

\begin{figure}
    \centering
    \includegraphics[width=\textwidth]{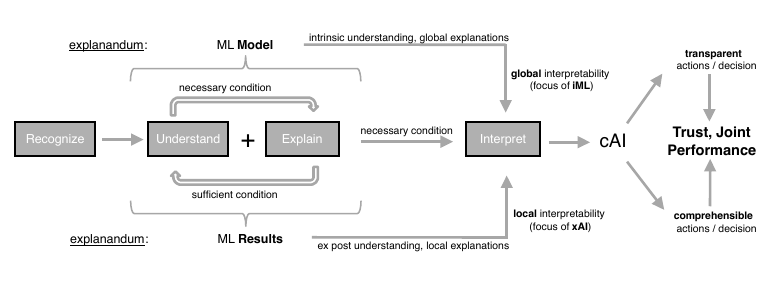}
    \caption{A framework for comprehensible artificial intelligence~\cite{bruckert2020next}.}
    \label{fig:cAI}
\end{figure}

Important work that is concerned with definitions for XAI can be found in, \forexample, \longcite{lipton_mythos_2018}, \longcite{adadi_peeking_2018}, and the work of \longcite{doshi-velez_rigorous_2017} who are often cited as base work for formalizing XAI terms (\cf~\forexample \cite{linardatos_explainable_2021}). Some definitions are taken from \longcite{bruckert2020next}, who present explanation as a process involving recognition, understanding and explicability respectively explainability, as well as interpretability.
The goal of the process is to achieve making an AI system's actions and decisions transparent as well as comprehensible.
An overview is given in Fig. \ref{fig:cAI}.

In the following we will assume we are given an AI system that should (partly) be explained to a human. The system encompasses one (or several) AI models and any pre- and post-processing. We use the following nomenclature:

\begin{description}[font=\normalfont\bfseries]
    \item[Understanding] is described as the human ability to recognize correlations, as well as the context of a problem and is a necessary precondition for explanations \cite{bruckert2020next}. The concept of understanding can be divided into mechanistic understanding (\emph{"How does something work?"}) and functional understanding (\emph{"What is its purpose?"})~\cite{paez2019pragmatic}.
    
    \item[Explicability] refers to making properties of an AI model inspectable~\cite{bruckert2020next}.
    
    \item[Explainability] goes one step further than \emph{explicability} and aims for making
    (a)~the context of an AI system's reasoning,
    (b)~the model, or
    (c)~the evidence for a decision output accessible,
    such that they can be \emph{understood} by a human~\cite{bruckert2020next}.
    
    \item[Transparency] is fulfilled by an AI model, if its algorithmic behaviour with respect to decision outputs or processes can be \emph{understood} by a human \emph{mechanistically}~\cite{paez2019pragmatic}.
    Transparency will be discussed more closely in \autoref{sec:taxonomy.problem.interpretability}.
    
    \item[Explaining] means utilizing \emph{explicability} or \emph{explainability} to allow a human to \emph{understand} a model and its purpose~\cite{bruckert2020next,paez2019pragmatic}.
    
    \item[Global explanations] \emph{explain} the model and its logic as a whole (\enquote{How was the conclusion derived?}).
        
    \item[Local explanations] \emph{explain} individual decisions or predictions of a model (\enquote{Why was this example classified as a car?}).
        
    \item[Interpretability] means that an AI model's decision can be \emph{explained globally} or \emph{locally} (with respect to \emph{mechanistic understanding}), and that the model's purpose can be \emph{understood} by a human actor \cite{paez2019pragmatic}(\idest \emph{functional understanding}).
    
    \item[Correctability] means that an AI system can be adapted by a human actor in a targeted manner in order to ensure correct decisions~\cite{kulesza2015principles,teso2019explanatory,schmid2020mutual}.
    Adaptation refers either to re-labelling of data~\cite{teso2019explanatory} or to changing of a model by constraining the learning process~\cite{schmid2020mutual}.
    
    \item[Interactivity] applies if one of the following is possible:
    (a)~interactive explanations, meaning a human actor can incrementally explore the internal working of a model and the reasons behind its decision outcome; or
    (b)~the human actor may adapt the AI system (\emph{correctability}).
    
    \item[Comprehensibility] relies, similar to \emph{interpretability}, on local and global \emph{explanations} and \emph{functional understanding}. Additionally, \emph{comprehensible} artificial intelligence fulfills \emph{interactivity}~\cite{bruckert2020next,schmid2020mutual}. 
    Both, \emph{interpretable} presentation and intervention are considered as important aspects for in depth \emph{understanding} and therefore preconditions to \emph{comprehensibility} (see also~\cite{gleicher_framework_2016}).
    
    \item[Human-AI system]
    is a system that contains both algorithmic components and a human actor,
    which have to cooperate to achieve a goal \cite{schmid2020mutual}.
    We here consider in specific \textbf{explanation systems}, \idest, such
    human-AI systems in which the cooperation involves \emph{explanations}
    about an algorithmic part of the system (the \emph{explanandum}) by
    an \emph{explanator} component, to the human interaction partner (the \emph{explainee})
    resulting in an action of the human~\cite{bruckert2020next}.
    
    \item[Explanandum]
    (\emph{what is to be explained}, \cf \autoref{sec:taxonomy.problem})
    refers to what is to be \emph{explained} in an \emph{explanation system}.
    This usually encompasses a model (\forexample, a deep neural network),
    We here also refer to an explanandum as the object of explanation.
    
    \item[Explanator]
    (\emph{the one that explains}, \cf \autoref{sec:taxonomy.explanator})
    is the \emph{explanation system} component providing \emph{explanations}.
    
    \item[Explainee]
    \emph{(the one to whom the explanandum is explained)}
    is the receiver of the \emph{explanations} in the \emph{explanation system}.
    Note that this often but not necessarily is a human. \emph{Explanations}
    may also be used \forexample, in multi-agent systems for communication
    between the agents and without a human in the loop in most of the information
    exchange scenarios.
    
    \item[Interpretable models] are defined as machine learning techniques that learn more structured representations, or that allow for tracing causal relationships. They are \emph{inherently interpretable} (\cf definition in  \autoref{sec:taxonomy.explanator}), \idest, no additional methods need to be applied to \emph{explain them}, unless the structured representations or relationship are too complex to be processed by a human actor at hand.
    
    \item[Interpretable machine learning (iML)] is the area of research concerned with the creation of \emph{interpretable} AI systems (\emph{interpretable models}).
    
    \item[Model induction] (also called model distillation, student-teacher approach, or reprojection~\cite{gleicher_framework_2016}) is a strategy that summarizes techniques which are used to infer an approximate \emph{explainable} model---the (\emph{explainable}) \emph{proxy} or \emph{surrogate model}---by observing the input-output behaviour of a model that is \emph{explained}.
    
    \item[Deep explanation] refers to combining deep learning with other methods in order to create hybrid systems that produce richer representations of what a deep neural network has learned, and that enable extraction of underlying semantic concepts \cite{gunning2019darpa}.

    \item[Comprehensible artificial intelligence (cAI)] is the result of a process that unites \emph{local interpretability} based on \emph{XAI} methods and \emph{global interpretability} with the help of \emph{iML} \cite{bruckert2020next}. The ultimate goal of such systems would be to reach \emph{ultra-strong machine learning}, where machine learning helps humans to improve in their tasks. For example, \cite{muggleton2018ultra} examined the \emph{comprehensibility} of programs learned with Inductive Logic Programming, and \cite{schmid2016does,schmid2020mutual} showed that the \emph{comprehensibility} of such programs could help laymen to \emph{understand} how and why a certain prediction was derived.
    
    \item[Explainable artificial intelligence (XAI)] is the area of research concerned with \emph{explaining} an AI system's decision.
    
\end{description}


\section{Approach to literature search}
\label{sec:approach}
The goals of this paper are to
(1)~provide a complete overview of relevant aspects or properties of XAI methods, and
(2) ease finding the right survey providing further details.
In order to achieve this, a systematic and broad literature analysis was conducted on papers in the time range of 2010 to 2021.
The target of this meta-survey are works that either contained reviews on XAI methods, or
considerations on XAI metrics and taxonomy aspects.

\subsection{Search}
Our search consisted of two iterations, one directly searching for work on XAI taxonomies,
and one more general search for general XAI surveys.

\paragraph*{Work on XAI taxonomies} 
The iteration for work on XAI taxonomies started with an initial pilot phase.
Here we identified common terms associated directly with XAI taxonomies
(for abbreviations both the abbreviation and the full expression must be considered):
\begin{itemize}
    \item machine learning terms: AI, DNN, Deep Learning, ML
    \item explainability terms: XAI, explain, interpret
    \item terms associated with taxonomies: taxonomy, framework, toolbox, guide
\end{itemize}
In the second search phase, we collected Google Scholar\footnote{\url{https://scholar.google.com}}
search results for combinations of these terms.
The main considered search terms are summarized in \autoref{tab:searchterms}.
For each search, the first 300 search results were scanned by the title for relevance to our search target.
Promising ones then were scanned by abstract.
Only work that we could access was finally included.
In the last phase, we scanned the works recursively for references to further relevant reviews.

\paragraph*{General XAI surveys} 
The second iteration collected search results for XAI surveys that
do not necessarily propose a taxonomy, but possibly implicitly use one.
For this, we again conducted the mentioned three search phases.
The search terms now were the more general ones
\enquote{XAI} and 
\enquote{XAI survey}.
These again were scanned first by title, then by abstract.
This resulted in a similar number of finally chosen and in-depth
assessed papers as the taxonomy search (not counting duplicate search results).

Lastly, we also included surveys and toolboxes that were additionally pointed out the reviewers.

\begin{table}[tbh]
    \centering
    \caption{Main used search phrases for the search for XAI taxonomies in the Google Scholar database with approximate number of matches}
    \label{tab:searchterms}
    \begin{tabularx}{\linewidth}{@{} r @{\hspace*{1em}}X @{}}
        \toprule
        \textbf{Matches}   & \textbf{Search phrase} \\\secseprule{2}
        $>300$    & explain AI taxonomy \\
        ca.\,$80$ & XAI taxonomy toolbox guide \\
        $>300$    & explainable AI taxonomy toolbox guide \\
        ca.\,$20$ & explain interpret AI artificial intelligence DNN Deep Learning ML machine learning taxonomy
framework toolbox guide XAI \\ 
        \bottomrule
    \end{tabularx}
\end{table}

\subsection{Categorization}
For the sub-selection and categorization, we considered the general focus, the length, level of detail, target audience, citation count per year, and recency.
\begin{description}[font=\normalfont\bfseries]
\item[General focus]
    Regarding the \emph{general focus}, we sorted the obtained reviews into three categories:
    \begin{description}
    \item[General XAI method collections (\autoref{sec:sos.general}):]
        Reviews that contain a broad collection of XAI methods without an explicit focus on a specific sub-field;
    \item[Domain-specific XAI method collections (\autoref{sec:sos.specific}):]
        Reviews that also contain a collection of XAI methods, but with a concrete focus on
        application domain (\forexample medicine),
        method technologies (\forexample inductive logic programming), or
        method traits (\forexample black-box methods);
    \item[Conceptual reviews (\autoref{sec:sos.conceptual}):]
       Reviews that do not contain or not focus on XAI method examples,
       but on conceptual aspects for the field of XAI like taxonomy or metrics;
    \item[Toolboxes (\autoref{sec:sos.toolbox}):]
       Summary of a complete toolbox with implementation.
    \end{description}
\item[Length]
    For the \emph{length} we considered the number of pages up to the bibliography, resulting in four categories (cf.~\autoref{fig:surveydist}):
    short (up to 6\,pages), medium (up to 15\,pages), long (up to 50\,pages), and very long (more than 50\,pages).
\item[Target audience]
    As \emph{target audiences} we considered three types:
    beginners in the field of XAI (potentially coming from a different domain),
    practitioners, and researchers.
    A review was associated with one or several target audiences,
    if it specifically targeted that audience, or
    if it was judged practically suitable or interesting for readers of that audience.
\item[Citation count per year]
    The \emph{citation count per year} of the surveys was used as a proxy for reception.
    It was collected from the popular citation databases
    Google Scholar,
    Semantic Scholar\footnote{\url{https://www.semanticscholar.org/}},
    OpenCitations\footnote{\url{https://opencitations.net/}},
    and NASA ADS\footnote{\url{https://ui.adsabs.harvard.edu/}}.
    The highest result (mostly google scholar) was chosen for comparison.
\item[Recency]
    was compared via the publication year.
\end{description}

Using these categorizations, some further works were ruled out for inclusion into the survey of surveys.
Excluded were reviews which:
would not sufficiently match our search target;
were too specific (\forexample comparison of only very few specific methods);
are very short (up to 4\,pages) and are covered by prior or successive work of the authors, or
are not often cited or not sufficiently focused.

\subsection{Results}
\begin{figure}
    \centering
    \begin{minipage}[t]{.5\linewidth}
        \centering
        \strut\\[-\baselineskip]%
        \includegraphics[width=0.95\linewidth]{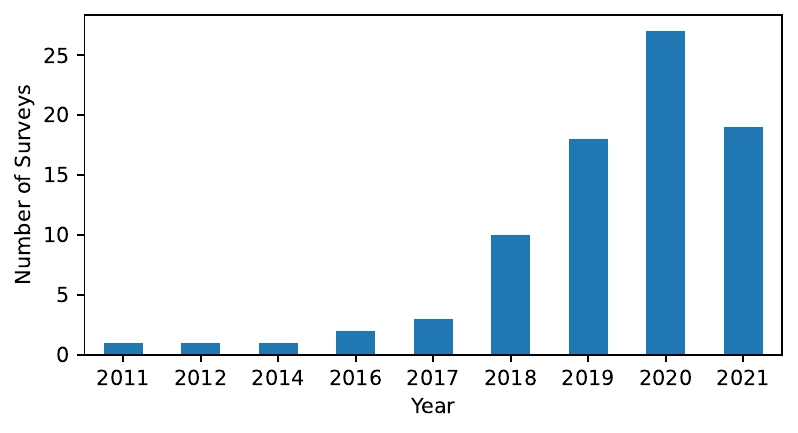}
    \end{minipage}%
    \begin{minipage}[t]{.5\linewidth}
        \centering
        \strut\\[-\baselineskip]%
        \includegraphics[width=0.95\linewidth]{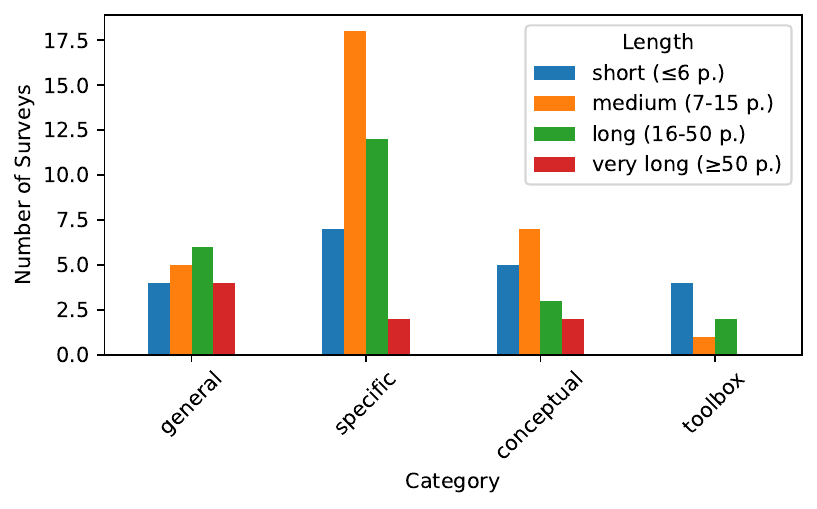}
    \end{minipage}
    \caption{Distribution of the reviewed surveys over the years 2010 to 2021 (\emph{left}), and distribution of the survey lengths by general focus category (\emph{right})}
    \label{fig:surveydist}
\end{figure}

We reviewed over 70 surveys on XAI with publication date up to beginning of 2021.
Most of them are from the years 2019 to 2021 (cf.~\autoref{fig:surveydist}),
which well fits the exponential increase in XAI methods that was observed so far \cite{arrieta_explainable_2020,linardatos_explainable_2021,adadi_peeking_2018,zhou_evaluating_2021}.
These were analysed for XAI method aspects, taxonomy structuring proposals,
and suitable example methods for each aspect.
In the following, a sub-selection of more than 50 surveys are detailed that were most cited or of general interest, as well as some examples toolboxes.

To exemplify the aspects of our proposed taxonomy, we selected again more than 50
concrete XAI methods that are reviewed in example sections for the corresponding XAI aspects.
The selection focused on high diversity and recency of the methods,
in order to establish a broad view on the XAI topic.
Finally, each of the methods were analysed on main taxonomy aspects, which is
summarized in \autoref{tab:methods.overview}.

\FloatBarrier
\section{A survey of surveys on XAI methods and aspects}
\label{sec:sos}

Along with the vastly growing demand and interest in XAI methods came an abundance of helpful reviews on the topic.
This encompasses overviews of all kinds of aspects, methods, and metrics, each with custom focus.
Given this battering amount of literature, it may be hard, especially for beginners and practitioners,
to find a survey suited to their needs.
The needs can encompass a specific focus, a desired level of detail and length, or others.
This chapter provides a short survey of XAI surveys that shall help a reader in finding appropriate material
to dig further into the topic of XAI.

The focus of the literature search lay on works that use, propose, or deduce a structured view on XAI methods and metrics.
Search and categorization criteria are detailed in \autoref{sec:approach}.
In the rest of this section, the more than 50 selected surveys are reviewed with respect to their
key focus points, level of detail, and their target audience.
While this survey of surveys for sure will not be complete, it should both give a good overview
on the breadth of XAI, as well as serve as a good starting point when looking for a deeper
dive into topics of the field.

Our results so far promise that many more helpful surveys on the topic of XAI will
come up within the next years (\cf \autoref{fig:surveydist}).
A key result of this structured meta-study will be presented later in \autoref{sec:taxonomy}:
The works are analyzed for taxonomy aspects of XAI methods, resulting in
the---to our knowledge---most complete taxonomy of XAI methods available so far.

\subsection{Conceptual reviews}\label{sec:sos.conceptual}
By now, many works have gathered and formulated important concepts related to XAI research.
We here roughly divided the available literature by their main focus:
broadly focused surveys, surveys from the perspectives of stakeholders and human-computer interaction,
and finally surveys with an explicit focus on XAI metrics.
An overview can be found in \autoref{tab:conceptualsurveys}.

\begin{table}
    \centering
    \caption{Overview on clusters of reviewed conceptual surveys on XAI}
    \label{tab:conceptualsurveys}
    \begin{tabulary}{\linewidth}{@{}l @{~~}L @{}}
        \toprule
        \textbf{Focus} & \\
        \secseprule{2}
        Broad &
            \longcite{gunning_xai_2019},
            \longcite{lipton_mythos_2018},
            \longcite{mueller_explanation_2019},
            \longcite{guidotti_principles_2021} \\
	    \midrule
	    Stakeholder perspective & 
	        \longcite{gleicher_framework_2016},
	        \longcite{langer_what_2021} \\
	    \midrule
        HCI perspective &
            \longcite{miller_explanation_2019},
            \longcite{chromik_taxonomy_2020},
            \longcite{ferreira_what_2020},
            \longcite{mueller_principles_2021} \\
        \midrule
	    XAI method evaluation &
	        \longcite{doshi-velez_rigorous_2017},
	        \longcite{zhou_evaluating_2021} \\
	    \bottomrule
    \end{tabulary}
\end{table}

\paragraph*{Broad conceptual surveys}
A very short, high-level, and beginner-friendly introduction to XAI can be found
in the work of \longcite{gunning_xai_2019}.
They derive four open challenges for the field: user-centric explanations, tradeoff between accuracy and interpretability, automated abstraction, and appropriate end-user trust. 
%
Regarding an XAI method taxonomy, a base for many later surveys was by
\longcite{lipton_mythos_2018}.
In this medium-length work, Lipton provides a small and high-level XAI taxonomy.
The focus, namely motivation and desiderata for interpretability, are held broad
and suitable for beginners and interdisciplinary discussion.
%
In contrast, the very broad and long DARPA report by \longcite{mueller_explanation_2019} provided later in 2019 is targeted at researchers of the field.
The goal of the report is to broadly capture all relevant developments and topics in the field of XAI.
This resulted in a detailed meta-study on state-of-the-literature, and
a detailed list of XAI method aspects and metrics (chapters 7 and 8).
%
More recently, \longcite{guidotti_principles_2021} illustrate some key dimensions to distinguish XAI approaches in a beginner-friendly book section. They present a broad collection of most common explanation types and state-of-the-art explanators respectively, and discuss their usability and applicability. This results in a conceptual review of taxonomy aspects and evaluation criteria.

\paragraph*{XAI from a stakeholder perspective}
The early work by \longcite{gleicher_framework_2016} highlights
the stakeholder perspective on XAI problems.
In the medium-length survey, Gleicher suggests a framework of general considerations
for practically tackling an explainability problem.
%
A similar focus is set by \longcite{langer_what_2021} in their long recent work from 2021.
They review in detail XAI from the point of view of satisfying stakeholder desiderata.
For this they collect standard goals of XAI problems and propagate that XAI design
choices should take into account all stakeholders in an interdisciplinary manner.

\paragraph*{XAI from a HCI perspective}
When humans interact with AI-driven machines, this human-machine-system
can benefit from explanations obtained by XAI.
Hence, there are by now several surveys concentrating on XAI
against the background of human-computer-interaction (HCI).
%
An important and well-received base work in this direction is that of
\longcite{miller_explanation_2019}.
He conducts a long, detailed, and extensive survey of work from philosophy, psychology, and
cognitive science regarding explanations and explainability.
The main conclusions for XAI research are:
(a)~(local) explanations should be understood contrastively,
    \idest, they should clarify why an action was taken instead of another;
(b)~explanations are selected in a biased manner,
    \idest, do not represent the complete causal chain but few selected causes;
(c)~causal links are more helpful to humans than probabilities and statistics; and
(d)~explanations are social
    in the sense that the background of the explanation receiver matters.
%
A much shorter paper was provided by \longcite{chromik_taxonomy_2020}. They rigorously develop a taxonomy for
evaluating black-box XAI methods with the help of human subjects.
Furthermore, they make some concrete suggestions for study design.
%
The related survey by \longcite{ferreira_what_2020} is slightly longer, and
may serve as an entry point to the topic for researchers.
They present a taxonomy of XAI that links with computer science and HCI communities,
with a structured and dense collection of many related reviews.
%
Very recent work on XAI metrics is provided by Müller \etal
in their 2021 paper \cite{mueller_principles_2021}.
They collect concrete and practical design principles for XAI in human-machine-systems.
Several relevant XAI metrics are recapitulated, and their broad collection of
related surveys may serve as an entry point to the research field of human-AI-systems.

\paragraph*{Surveys with a focus on evaluation}
Continuing on the HCI perspective, a hot topic in the field of XAI are metrics for measuring the quality
of explanations for human receivers.
An early and base work on XAI metric categorization is by \longcite{doshi-velez_rigorous_2017}. The medium-length survey collects latent dimensions of
interpretability with recommendations on how to choose and evaluate an XAI method.
Their taxonomy for XAI metrics is adapted by us, classifying metrics into human grounded, functionally grounded, and application grounded ones (\cf \autoref{sec:taxonomy.metrics}).
%
A full focus on metrics for evaluating XAI methods is set in the
recent work by \longcite{zhou_evaluating_2021}.
This medium-length, detailed meta-study reviews diverse XAI metrics,
aligned with shallow taxonomies both for methods and metrics.

\FloatBarrier
\subsection{Broad method collections}\label{sec:sos.general}
There by now exists an abundance of surveys containing broad collections of XAI methods,
serving as helpful starting points for finding the right method.
The following shortly highlights surveys that feature a generally broad focus
(for specifically focused surveys see \autoref{sec:sos.specific}).
This summary shall help to find a good starting point for diving deeper into methods.
Hence, the surveys are first sorted by their length (as a proxy for the amount of information),
and then by their reception (citations per year).
The latter was found to strongly correlate with age.

\begin{table}
    \centering
    \caption{
        Overview on focus points/specialties and audience (Aud.) of discussed general XAI method collections.
        Sorted first by length, then reception (citations per year).
        Detail is manually ranked in values from 1 (short discussion per method) to 5 (detailed discussion per method).
        The audience can be beginner (B), practitioner (P), or researcher (R).
    }
    \label{tab:generalsurveys}
    \footnotesize
    \begin{tabulary}{\linewidth}{@{} l @{~}l@{~~} L@{}c @{}}
    \toprule
    		&	\textbf{Aud.}	&	\textbf{Focus / Specialty}	&	\textbf{Detail}
\\\secseprule{4}\multicolumn{4}{@{}l}{\textbf{Short}}														
	\\ \midrule \longcite{dosilovic_explainable_2018}	&	B, R	&	XAI for supervised learning	&	1
	\\ \midrule \longcite{biran_explanation_2017}	&	R	&	Local explanations and interpretable models	&	1
	\\ \midrule \longcite{benchekroun_need_2020}	&	B, P	&	Industry point of view	&	5
\\\secseprule{4}\multicolumn{4}{@{}l}{\textbf{Medium}}														
	\\\midrule \longcite{gilpin_explaining_2018}	&	R	&	XAI research aspects	&	1
	\\ \midrule	\longcite{du_techniques_2019}	&	B	&	Perturbation based attention visualization	&	3
	\\ \midrule \longcite{murdoch_definitions_2019}	&	B, P	&	Predictive, descriptive, relevant desiderata for explanations, practical examples	&	5
	\\ \midrule \longcite{goebel_explainable_2018}	&		&	Need for XAI	&	3
	\\ \midrule \longcite{islam_explainable_2021}	&	B, R	&	Demonstration of XAI methods in case-study (credit scoring)	&	5
\\\secseprule{4}\multicolumn{4}{@{}l}{\textbf{Long}}														
	\\ \midrule \longcite{adadi_peeking_2018}	&	P	&	XAI research landscape (methods, metrics, cognitive aspects, human-machine-systems)	&	1
	\\ \midrule \longcite{carvalho_machine_2019}	&	B, R	&	Aspects associated with XAI (esp.\ motivation, properties, desirables)	&	2
	\\ \midrule \longcite{xie_explainable_2020}	&	B	&	Explanation of DNNs	&	3
	\\ \midrule \longcite{das_opportunities_2020}	&	R	&	Visualization methods for visual DNNs	&	3
	\\ \midrule \longcite{linardatos_explainable_2021}	&	P, R	&	Technical, with source code	&	3
	\\ \midrule \longcite{bodria_benchmarking_2021}	&	B, P	&	Comparative studies of XAI methods	&	4
\\\secseprule{4}\multicolumn{4}{@{}l}{\textbf{Very long}}														
	\\ \midrule \longcite{molnar_interpretable_2020}	&	B	&	Fundamental concepts of interpretability	&	5
	\\ \midrule \longcite{arrieta_explainable_2020}	&	P, R	&	Notion of responsible AI: Terminology, broad collection of examples	&	1
	\\ \midrule \longcite{burkart_survey_2021}	&	R	&	Supervised machine learning; linear, rule-based methods and decision trees, data analytics and ontologies	&	3
	\\ \midrule \longcite{vilone_explainable_2020}	&	R	&	Systematic and broad review of XAI surveys, theory, methods, and method evaluation	&	3
	\\\bottomrule
    \end{tabulary}
\end{table}

\paragraph*{Short surveys}
A short overview mostly on explanations of supervised learning approaches
was provided by Došilović \etal in 2018	\cite{dosilovic_explainable_2018}.
This quite short but broad and beginner-friendly introductory survey
shortly covers diverse approaches and open challenges towards XAI.
%
Slightly earlier in 2017, Biran and Cotton published their XAI survey \cite{biran_explanation_2017}.
This is a short and early collection of explainability concepts. Their general focus
lies on explanations of single predictions, and diverse model types like rule-based
systems and Bayesian networks, which are each shortly discussed.
%
Most recently, in 2020 Benchekroun \etal collected and presented XAI methods
from an industry point of view \cite{benchekroun_need_2020}.
They present a preliminary taxonomy that includes pre-modelling explainability as
an approach to link knowledge about data with knowledge about the used model and its results.
Regarding the industry perspective, they specifically motivate standardization.

\paragraph*{Medium-length surveys}
An important and very well received earlier work on XAI method collections was provided
by \longcite{gilpin_explaining_2018}.
Their extensive survey includes very different kinds of XAI methods, including, e.g., rule extraction.
In addition, for researchers they provide references to further more specialized surveys in the field.
It is very similar to the long survey by \longcite{adadi_peeking_2018}, only shortened by skipping detail on the methods.
%
More detail, but a slightly more specialized focus, is provided in the survey by \longcite{du_techniques_2019}.
The beginner-friendly high-level introduction to XAI features few, in detail discussed examples.
These concentrate on perturbation based attention visualization.
%
Similarly, the examples in \cite{murdoch_definitions_2019} also mostly focus on
visual domains and explanations. This review by Murdoch \etal in 2019 is also beginner-friendly,
and prefers detail over covering a high number of methods. The examples are embedded into a comprehensive short
introductory review of key categories and directions in XAI.
Their main focus is on the proposal of three simple practical desiderata for explanations:
The model to explain should be predictive (predictive accuracy),
the explanations should be faithful to the model (descriptive accuracy), and
the information presented by the explanations should be relevant to the receiver.
%
Slightly earlier, \longcite{goebel_explainable_2018} concentrate in their work more on multimodal explanations and question-answering systems.
This survey contains a high-level review of the need for XAI, and discussion of some
exemplary state-of-the-art methods.
%
A very recent and beginner-friendly XAI method review is by \longcite{islam_explainable_2021}. Their in-depth discussion of example methods is aligned with
a shallow taxonomy, and many examples are practically demonstrated in a common simple case study.
Additionally, they present a short meta-study of XAI surveys, and a collection of future perspectives for XAI.
The latter includes formalization, XAI applications for fair and accountable ML, XAI for human-machine-systems,
and more interdisciplinary cooperation in XAI research.

\paragraph*{Long surveys}
A well-received lengthy and extensive XAI literature survey was conducted by
\longcite{adadi_peeking_2018}.
They reviewed and shortly discuss 381 papers related to XAI to provide a holistic
view on the XAI research landscape at that time. This includes methods, metrics,
cognitive aspects, and aspects related to human-machine-systems.
%
Another well-received, but more recent, slightly less extensive and more
beginner-friendly survey is that by \longcite{carvalho_machine_2019}.
They collect and discuss in detail different aspects of XAI, especially motivation,
properties, desirables, and metrics. Each aspect is accompanied by some examples.
%
Similarly beginner-friendly is the introductory work of \longcite{xie_explainable_2020}.
Their field guide explicitly targets newcomers to the field with a general
introduction to XAI, and a wide variety of examples of standard methods,
mostly for explanations of DNNs.
%
A more formal introduction is provided by \longcite{das_opportunities_2020} in their survey.
They collect formal definitions of XAI related terms, and develop a shallow taxonomy.
The focus of the examples is on visual local and global explanations of DNNs based on
(model-specific) backpropagation or (model-agnostic) perturbation-based methods.
More recent and much broader is the survey of \longcite{linardatos_explainable_2021}.
It provides an extensive technical collection and review of XAI methods with code
and toolbox references, which makes it specifically interesting for practitioners.
%
Similarly and in the same year, \longcite{bodria_benchmarking_2021} review in detail more than 60 XAI methods for visual, textual, and tabular data models.
These are selected to be most recent and widely used, and cover a broad range of explanation types. Several comparative benchmarks of methods are included, as well as a short review of toolboxes.

\paragraph*{Very long surveys}
By now there are a couple of surveys available that aim to give a broad, rigorous, and
in-depth introduction to XAI.
%
A first work in this direction is the regularly updated book on XAI by
\longcite{molnar_interpretable_2020}, first published in 2017\footnote{\url{https://github.com/christophM/interpretable-ml-book/tree/v0.1}}.
This book is targeted at beginners, and gives a basic and detailed introduction on
interpretability methods, including many transparent and many model-agnostic ones.
The focus lies more on fundamental concepts of interpretability and
detail on standard methods than on the amount of discussed methods.
%
Meanwhile, \longcite{arrieta_explainable_2020} put up a very long and broad collection of XAI methods
in their 2020 review.
The well-received survey can also be considered a base work of state-of-the-art XAI,
as they introduce the notion of \emph{responsible AI}, \idest, development of AI models
respecting fairness, model explainability, and accountability.
This is also the focus of their work, in which they provide terminology,
a broad but less detailed selection of example methods, and practical discussion
for responsible AI.
%
A very recent extensive XAI method survey is that of \longcite{burkart_survey_2021}. They review in moderate detail explainability
methods, primarily for classification and regression in supervised machine learning.
Specifically, they include many rule-based and decision-tree based explanation methods,
as well as aspects on data analysis and ontologies for formalizing input domains.
This is preceded by a deep collection of many general XAI aspects.
%
Finally and a little earlier, \longcite{vilone_explainable_2020} published in 2020 an
equally extensive systematic literature survey on XAI in general.
The systematic literature search was similar to ours, only with a different focus.
They include a broad meta-study of reviews, as well as reviews and discussion of works on
XAI theory, methods, and method evaluations.

\FloatBarrier
\subsection{Method collections with specific focus}\label{sec:sos.specific}
Besides the many broad method collections, there by now are numerous ones specifically
concentrating on an application domain, specific input or task types, certain surrogate model types,
or other traits of XAI methods.
We here manually clustered surveys by similarity of their main focus points.
An overview on the resulting clusters is given in \autoref{tab:specificsurveys}.

\begin{table}
    \centering
    \footnotesize
    \caption{Overview on surveys clusters reviewed \autoref{sec:sos.specific} with restricted (restr.) focus}
    \label{tab:specificsurveys}
    \begin{tabularx}{\linewidth}{@{}>{\raggedright}p{6em} >{\raggedright}p{8.5em} X @{}}
        \toprule
        \textbf{Restr. by:} & \textbf{Restr. to:} & \\
        \secseprule{3}
        Application domain
                      & NLP & \longcite{danilevsky_survey_2020} \\\cmidrule{2-3}
	                  & Medicine & \longcite{singh_explainable_2020,tjoa_survey_2020} \\\cmidrule{2-3}
	                  & Recommendation systems & \longcite{nunes_systematic_2017,zhang_explainable_2020} \\
	    \midrule
	    Application type
	                  & Interactive ML & \longcite{anjomshoae_explainable_2019,baniecki_grammar_2020,amershi_power_2014} \\
	    \midrule
        Task          
                      & Visual tasks  &  \longcite{samek_explainable_2019a,samek_explainable_2019,nguyen_understanding_2019,ancona_gradientbased_2019,alber_software_2019,li_quantitative_2020,zhang_visual_2018}\\\cmidrule{2-3}
                      & Reinforcement ML &  \longcite{puiutta_explainable_2020,heuillet_explainability_2021} \\
        \midrule
	    Explanator output type
	                  & Rule-based XAI & \longcite{cropper_turning_2020,vassiliades_argumentation_2021,hailesilassie_rule_2016,calegari_integration_2020} \\\cmidrule{2-3}
	                  & Counterfactual explanations & \longcite{byrne_counterfactuals_2019,artelt_computation_2019,verma_counterfactual_2020,keane_if_2021,mazzine_framework_2021,karimi_survey_2021,stepin_survey_2021} \\
	    \midrule
	    Other XAI method traits
	                  & Model-agnostic methods &  \longcite{guidotti_survey_2018} \\
        \bottomrule
    \end{tabularx}
\end{table}

\paragraph*{XAI for specific application domains}
Some method surveys focus on concrete practical application domains.
%
One is by \longcite{danilevsky_survey_2020}, who survey
XAI methods for \emph{natural language processing} (NLP). This includes a taxonomy, a
review of several metrics, and a dense collection of XAI methods.
%
Another important application domain is the \emph{medical domain}.
For example, \longcite{singh_explainable_2020} provided a survey and taxonomy of XAI methods for image classification with a focus on medical applications.
%
A slightly broader focus on general XAI methods for medical applications was selected by \longcite{tjoa_survey_2020} in the same year.
Their long review shortly discusses more than sixty methods, and sorts them into a shallow taxonomy.
%
Another domain sparking needs for XAI is that of \emph{recommendation systems},
\forexample, in online shops.
%
An example here is the long, detailed, and practically oriented survey by \longcite{nunes_systematic_2017}. Amongst others, they present a detailed taxonomy
of XAI methods for recommendation systems (\cf \cite[Fig.\,11]{nunes_systematic_2017}).
%
A similar, but even more extensive and lengthy survey was provided by
\longcite{zhang_explainable_2020}.
They generally review recommendation systems also in a practically oriented manner,
and provide a good overview on models that are deemed explainable.

\paragraph*{XAI for interactive ML applications}
Several studies concentrate on XAI to realize interactive machine learning.
For example, \longcite{anjomshoae_explainable_2019} reviewed explanation generation, communication and
evaluation for autonomous agents and human-robot interaction.
%
\longcite{baniecki_grammar_2020} instead directly concentrated on interactive machine learning
in their medium-length survey on the topic.
They present challenges in explanation, traits to overcome these, as well as a taxonomy
for interactive explanatory model analysis.
%
The longer but earlier review by \longcite{amershi_power_2014}
more concentrates on practical case studies and research challenges.
They motivate incremental, interactive and practical human-centered XAI methods.

\paragraph*{XAI for visual tasks}
Some of the earlier milestones for the current field of XAI were methods
to explain input importance for models with image inputs \cite{das_opportunities_2020},
such as LIME~\cite{ribeiro_why_2016} and LRP~\cite{bach_pixelwise_2015}.
Research on interpretability and explainability of models for visual tasks is still
very active, as several surveys with this focus show.
%
One collection of both methods and method surveys with a focus on visual explainability
is the book edited by \longcite{samek_explainable_2019a}.
This includes the following surveys:
\begin{itemize}
    \item \longcite{samek_explainable_2019}:
    A short introductory survey on visual explainable AI for researchers, giving an overview on important developments;
    \item \longcite{nguyen_understanding_2019}:
    A survey specifically on feature visualization methods.
    These are methods to find prototypical input patterns that most activate parts of a DNN.
    The survey includes a mathematical perspective on the topic and a practical overview on applications.
    \item \longcite{ancona_gradientbased_2019}:
    A detailed survey on gradient-based methods to find attribution of inputs to outputs; and
    \item \longcite{alber_software_2019}:
    A detailed collection of implementation considerations regarding different methods
    for highlighting input attribution.
    The review includes code snippets for the TensorFlow deep learning framework.
\end{itemize}
%
Similar to \longcite{ancona_gradientbased_2019},
\longcite{li_quantitative_2020} focus on XAI methods to obtain
heatmaps as visual explanations. They in detail discuss seven examples of methods,
and conduct an experimental comparative study with respect to five specialized metrics.
%
Another survey focusing on visual interpretability is the earlier work
by \longcite{zhang_visual_2018}. This well-received medium-length
survey specializes on visual explanation methods for convolutional neural networks.

\paragraph*{XAI for reinforcement learning tasks}
Just as for visual tasks, there are some studies specifically focusing on
explanations in tasks solved by reinforcement learning.
One is that by \longcite{puiutta_explainable_2020}.
This medium long to lengthy review provides a short taxonomy on XAI methods for reinforcement learning.
It reviews more than 16 methods specific to reinforcement learning in a beginner-friendly way.
%
A comparable and more recent, but slightly longer, more extensive, and more technical
survey on the topic is by \longcite{heuillet_explainability_2021}.

\paragraph*{XAI methods based on rules}
One type of explanation outputs is that of (formal) symbolic rules.
Both generation of interpretable rule-based models, as well as extraction of
approximate rule sets from less interpretable models have a long history.
We here collect some more recent reviews on these topics.
%
One is the historical review by \longcite{cropper_turning_2020} on the developments in Inductive Logic Programming. Inductive logic programming summarizes methods to automatically construct
rule-sets for solving a task given some formal background knowledge and few examples.
The mentioned short survey aims to look back at the last 30 years of development in the field,
and serve as a good starting point for beginners.
%
On the side of inherently interpretable rule-based models, \longcite{vassiliades_argumentation_2021} recently 
reviewed argumentation frameworks in detail.
An argumentation framework provides an interpretable logical argumentation line that formally
deduces a statement from (potentially incomplete) logical background knowledge.
This can, \forexample, be used to find the most promising statement from some choices,
and, hence, as an explainable (potentially interactive) model on symbolic data.
The long, detailed, and extensive survey formally introduces standard argumentation frameworks,
reviews existing methods and applications, and promotes argumentation frameworks as
promising interpretable models.
%
While the previous surveys concentrate on directly training inherently
interpretable models consisting of rules, \longcite{hailesilassie_rule_2016} shortly reviewed rule extraction methods.
Rule extraction aims to approximate a trained model with symbolic rule sets or decision trees.
These methods have a long history but faced their limits when applied to large-sized models
like state-of-the-art neural networks, as discussed in the survey.
%
A more recent survey that covers both rule extraction as well as integration of symbolic
knowledge into learning processes is the review by
\longcite{calegari_integration_2020}.
Their long and in-depth overview covers the main symbolic/sub-symbolic integration
techniques for XAI, including rule extraction methods for some steps.

\paragraph*{Counterfactual and contrastive explanations}
While the previous surveys were mostly concentrated on a specific type of task,
there are also some that are restricted to certain types of explainability methods.
One rising category is that of contrastive and counterfactual explanations (also \emph{counterfactuals}).
The goal of these is to explain for an instance \enquote{why was the output $P$ rather than $Q$?} \cite{stepin_survey_2021}, and, in particular for counterfactual explanations, how input features can be changed to achieve $Q$ \cite{mazzine_framework_2021}. To do this, one or several other instances are provided to the user that produce the desired output.
%
One is the short survey by \longcite{byrne_counterfactuals_2019}.
This reviews specifically counterfactual explanations with respect to evidence from human reasoning.
The focus here lies on additive and subtractive counterfactual scenarios.
%
Also as early as 2019, \longcite{artelt_computation_2019} provide a beginner-friendly review on model-specific counterfactual explanation methods. They consider a variety of standard ML models, and provide detailed mathematical background for each of them.
%
A more broad survey on counterfactuals was conducted by \longcite{verma_counterfactual_2020}. The mid-length survey reviews 39 methods and discusses common desiderata, a taxonomy, evaluation criteria, and open challenges for this XAI subtopic. In particular, they suggest the following desiderata:
counterfactual examples should be valid inputs that are similar to the training data;
the example should be as similar as possible to the original (proximity), while changing as few features as possible (sparsity);
feature changes should be actionable, \idest, the explainee should be able to achieve them (\forexample, increase age, not decrease);
and they should be \emph{causal}, acknowledging known causal relationships in the model.
Later studies confirm these desirables.
%
In particular, \longcite{keane_if_2021} review in total 100 methods with respect to common motivations for counterfactual explanations and typical shortcomings thereof, in order to guide researchers. In their short survey, they find that better psychological grounding of counterfactuals as well as their evaluation is required, in specific for validity and feature selection. Also, methods up to that point in time are often missing user studies and comparative tests.
%
A closer look at how to tackle comparative studies was taken by \longcite{mazzine_framework_2021} in their extensive survey.
They benchmarked open source implementations of 10 strategies for counterfactual generation for DNNs on 22 different tabular datasets. The controlled experimental environment may serve as a role model for researchers for future evaluations.
%
In contrast to functional aspects of counterfactual methods, \longcite{karimi_survey_2021} in parallel focused on the use-case perspective: The mid-length survey highlights the impact of the mentioned desiderata on the use-case of algorithmic recourse.
Algorithmic recourse here means to provide explanations and actionable recommendations for individuals who encountered an unfavorable treatment by an automated decision-making system.
%
Lastly, the broad and extensive survey by \longcite{stepin_survey_2021} by unites research perspectives on conceptual and methodological work for advanced researchers.
They rigorously survey 113 studies on counterfactual and contrastive explanations reaching back as far as the 1970s. The reader is provided with a review of terms and definitions used throughout the primary literature, as well as a detailed taxonomy. The obtained conceptual insights are then related to the methodological approaches.

\paragraph*{Model-agnostic XAI methods}
A very well received, long and extensive survey for model-agnostic XAI methods on tabular data is by
\longcite{guidotti_survey_2018}.
Besides the method review, they also develop a formal approach to define XAI use-cases
that is especially useful for practitioners.

\FloatBarrier
\subsection{Toolboxes}\label{sec:sos.toolbox}

A single XAI method often does not do the full job of making all relevant
model aspects clear to the explainee.
Hence, toolboxes have become usual that implement more than one explainability
method in a single library with a common interface.
For detailed lists of available toolboxes the reader is referred to, e.g.,
the related work in \cite[Tab.\,1]{arya_one_2019}, the repository links in
\cite[Tabs.\,A.1, A.2]{linardatos_explainable_2021}, and the toolbox review in \cite[Sec.\,7]{bodria_benchmarking_2021}.
For implementation considerations in the case of visual interpretability
the review \cite{alber_software_2019} is a suitable read.
We here provide some examples of publications presenting toolboxes published since 2019 that we analyzed for taxonomy aspects, summarized in \autoref{tab:toolboxes}.

\begin{table}
    \centering
    \caption{Overview on the sub-selection of papers introducing explainability toolboxes that were considered in \autoref{sec:sos.toolbox} with their respective code repository;
    more extensive overviews can be found, e.g., in
    \cite[Tab.\,1]{arya_one_2019}, \cite[Tabs.\,A.1, A.2]{linardatos_explainable_2021}, and \cite[Sec.\,7]{bodria_benchmarking_2021}.
    }
    \label{tab:toolboxes}
    \begin{tabular}{@{}l l l @{}}
        \toprule
        \textbf{Toolbox} & \textbf{Publication} 
        & \textbf{Code repository} \\
        \secseprule{3}
        Skater & \longcite{choudhary_interpreting_2018} 
        & \url{https://github.com/oracle/Skater} \\
        InterpretML & \longcite{nori_interpretml_2019} 
        & \url{https://github.com/interpretml/interpret} \\
        iNNvestigate & \longcite{alber_innvestigate_2019} 
        & \url{https://github.com/albermax/innvestigate} \\
        AI Fairness 360 & \longcite{arya_one_2019} 
        & \url{https://github.com/Trusted-AI/AIF360} \\
        explAIner & \longcite{spinner_explainer_2020} 
        & \url{https://github.com/dbvis-ukon/explainer} \\
        FAT Forensics & \longcite{sokol_fat_2020} 
        & \url{https://github.com/fat-forensics/fat-forensics} \\
        Alibi & \longcite{klaise_alibi_2021} 
        & \url{https://github.com/SeldonIO/alibi} \\
        \bottomrule
    \end{tabular}
\end{table}

%
Already in 2018, first toolboxes like Skater \cite{choudhary_interpreting_2018} were available for several XAI tasks. Skater, in specific, provides in total seven XAI methods for both global and local explanation of different kinds of trained models, including DNNs for visual and textual inputs.
%
The beginner-friendly Microsoft toolbox InterpretML~\cite{nori_interpretml_2019}
implements five model-agnostic and four transparent XAI methods
that are shortly introduced in the paper.
%
Another toolbox from that year is iNNvestigate by \longcite{alber_innvestigate_2019}.
They specifically concentrate on some standard post-hoc heatmapping methods for visual explanation.
%
\longcite{arya_one_2019} presented the IBM AI explainability 360 toolbox
with 8 diverse XAI methods in 2019.
The implemented workflow follows a proposed practical, tree-like taxonomy of XAI methods.
Implemented methods cover a broad range of explainability needs, including
explainability of data, inherently interpretable models, and post-hoc explainability
both globally and locally.
%
More recently, \longcite{spinner_explainer_2020} presented their toolbox explAIner.
This is realized as a plugin to the existing TensorBoard\footnote{TensorBoard toolkit: \url{https://www.tensorflow.org/tensorboard}} toolkit for the TensorFlow deep learning framework.
The many post-hoc DNN explanation and analytics tools are aligned with the suggested
pipeline phases of model understanding, diagnosis, and refinement.
Target users are both researchers, practitioners, and beginners.
%
In 2020 also the FAT Forensics (for Fairness, Accountability, Transparency) toolbox was published \cite{sokol_fat_2020}. Based on scikit-learn, several applications towards model and data quality checks and local and global explainability are implemented.
The focus is on black-box methods, providing a generic interface for two use-cases: research on new fairness metrics (for researchers) and monitoring of ML models during pre-production (for practitioners).
At the time of writing, methods for image and tabular data are supported.
%
The most recent toolbox considered here is the Alibi explainability kit by \longcite{klaise_alibi_2021}. It features nine diverse, mostly black-box XAI methods for classification and regression. This includes both local and (for tabular data) global state-of-the-art analysis methods for image or tabular input data, with some additionally supporting textual inputs.
Alibi targets practitioners, aiming to be an extensively tested, production-ready, scalable, and easy to integrate toolbox for explanation of machine learning models.


\section{Taxonomy} 
\label{sec:taxonomy}

\begin{figure}
    \centering
    \includegraphics[scale=\thetaxonomygraphicscale]{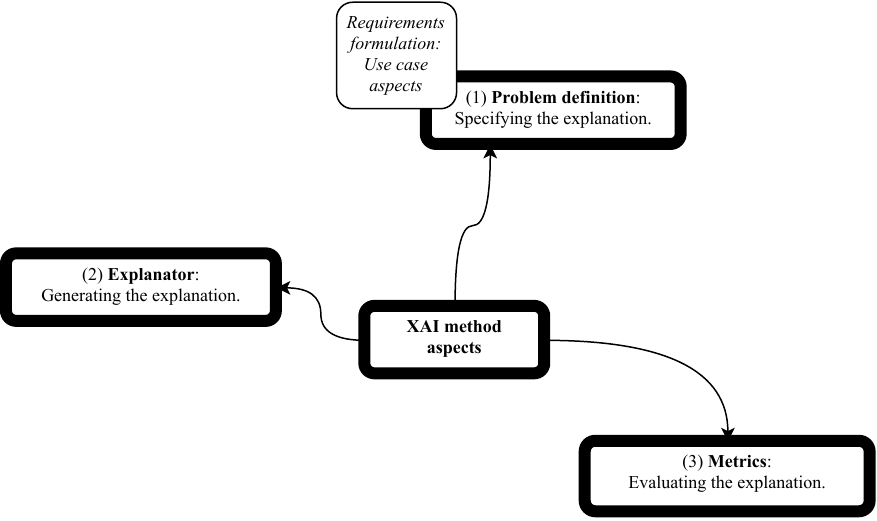}
    \caption{Overview of the top-level categorization of taxonomy aspects explained in \autoref{sec:taxonomy}. Find a visualization of the complete taxonomy in \autoref{fig:taxonomy} on page \pageref{fig:taxonomy}.}
    \label{fig:taxonomy.base}
\end{figure}

\begin{landscape}
\begin{figure}
    \raggedright
    \includegraphics[scale=\thetaxonomygraphicscale]{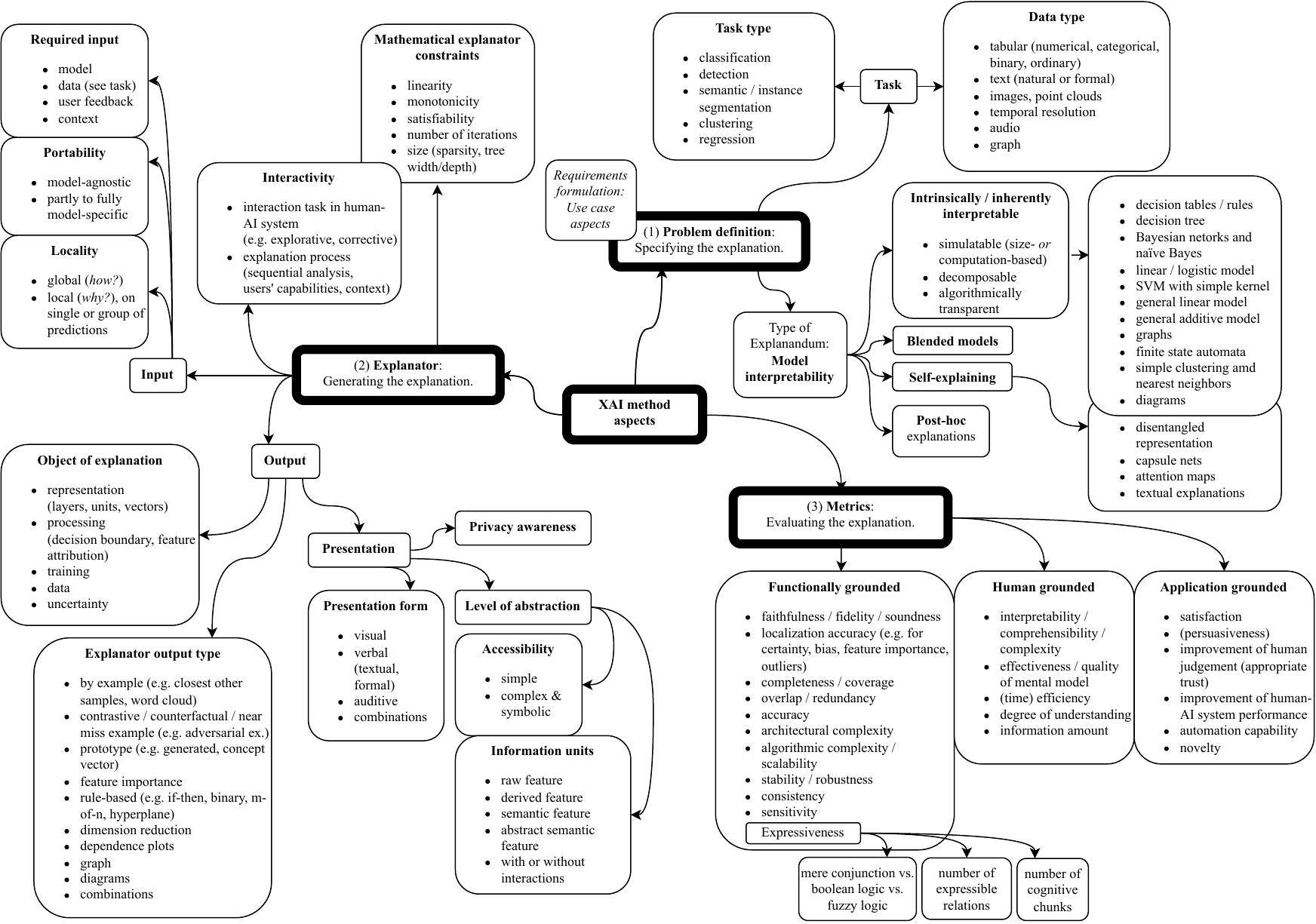}
    \caption{Overview of the complete taxonomy that is detailed in \autoref{sec:taxonomy}}
    \label{fig:taxonomy}
\end{figure}
\end{landscape}

This section establishes a taxonomy of XAI methods that unites terms and notions from the literature and that helps to differentiate and evaluate XAI methods.
More than 70 surveys that have been selected in the course of our literature search (\cf \autoref{sec:approach}), including the ones discussed in \autoref{sec:sos}, which were analyzed for such terms.
We then identified synonymous terms.
Finally, we sub-structured the notions systematically according to practical considerations in order to provide a complete picture of the state-of-the-art XAI method and evaluation aspects.
This section details the found notions and synonyms, and our proposed structure thereof, which defines the outline.
An overview of the root-level structure is provided in \autoref{fig:taxonomy.base}, and a complete overview of the final structure in \autoref{fig:taxonomy}.
The sub-trees for the first-level categories can be found at the beginning of each section.

At the root level, we propose to categorize in a \emph{procedural manner} according to the steps for building an explanation system.
The following gives a high-level overview of the first two levels of the taxonomy structure and the outline of this chapter
(\cf overview in \autoref{fig:taxonomy.base}):
\begin{enumerate}
    \item \textbf{Problem definition} (\autoref{sec:taxonomy.problem}):
    One usually should start with the \emph{problem definition}. 
    This encompasses
    \begin{itemize}
        \item traits of the \emph{task}, and
        \item the \emph{explanandum} (precisely: the \emph{interpretability} of the explanandum).
    \end{itemize}
    \item \textbf{Explanator properties} (\autoref{sec:taxonomy.explanator}):
    Then, we detail the \emph{explanator properties} (\autoref{sec:taxonomy.explanator}), which we functionally divided into properties of
    \begin{itemize}
        \item \emph{input},
        \item \emph{output},
        \item \emph{interactivity} with the user, and
        \item any further \emph{formal constraints} posed on the explanator.
    \end{itemize}
    \item \textbf{Metrics} (\autoref{sec:taxonomy.metrics}):
    Lastly, we discuss different \emph{metrics} (\autoref{sec:taxonomy.metrics})
    that can be applied to explanation systems in order to evaluate their qualities. Following \longcite{doshi-velez_rigorous_2017}, these are divided by their dependence on subjective human evaluation and the application into:
    \begin{itemize}
        \item \emph{functionally-grounded} metrics (independent of human judgment),
        \item \emph{human-grounded} metrics (subjective judgment required), and
        \item \emph{application-grounded} (full human-AI-system required).
    \end{itemize}
\end{enumerate}
The presented aspects are illustrated by selected example methods (marked in {\color{gray}gray}).
The selection of example methods is by no means complete. Rather it intends to give
an impression about the wide range of the topic and how to apply our taxonomy to
both some well-known and less known but interesting methods.

\paragraph*{On requirements derivation}
Note that from a procedural perspective, the first step when designing an explanation system should be to
determine the use-case-specific \emph{requirements} (as part of the problem definition).
The requirements can be derived from all the taxonomy aspects that are collected in this review. This gives rise to a similar sub-structuring as the procedural one shown above:
\begin{itemize}
    \item Both explanandum and explanator must match the \emph{task},
    \item the explanandum should be chosen to match the (inherent) \emph{interpretability} needs, and
    \item the explanator must fulfill any other (functional and architectural) \emph{explanator constraints} (see \autoref{sec:taxonomy.explanator}), as well as
    \item any \emph{metric target values}.
\end{itemize}
This should be motivated by the actual goal or desiderata of the explanation.
These can be, \forexample, verifiability of properties like fairness, safety, and security,
knowledge discovery, promotion of user adoption respectively trust,
or many more. An extensive list of desiderata can be found in the work of \longcite{langer_what_2021}.
As noted in \autoref{sec:relatedwork}, a detailed collection of XAI needs, desiderata, and typical use-cases is out of the scope of this work.
Instead, our collection of taxonomy aspects shall serve as a starting point for effective and complete use-case analysis and requirements derivation.


\subsection{Problem definition}\label{sec:taxonomy.problem}
The following aspects consider the concretion of the explainability problem.
Apart from the use-case analysis, which is skipped in this work,
details on the following two aspects must be clear:
\begin{itemize}
    \item \textbf{\hyperref[sec:taxonomy.problem.task]{Task}}:
        the \emph{task} that is to be explained must be clear, and
    \item \textbf{\hyperref[sec:taxonomy.problem.interpretability]{Model interpretability}}:
        the solution used for the task, meaning the \emph{type of explanandum}.
        For explainability purposes, the level of \emph{interpretability} of the explanandum model is the relevant point here.
\end{itemize}

\begin{figure}
    \centering
    \includegraphics[scale=\thetaxonomygraphicscale]{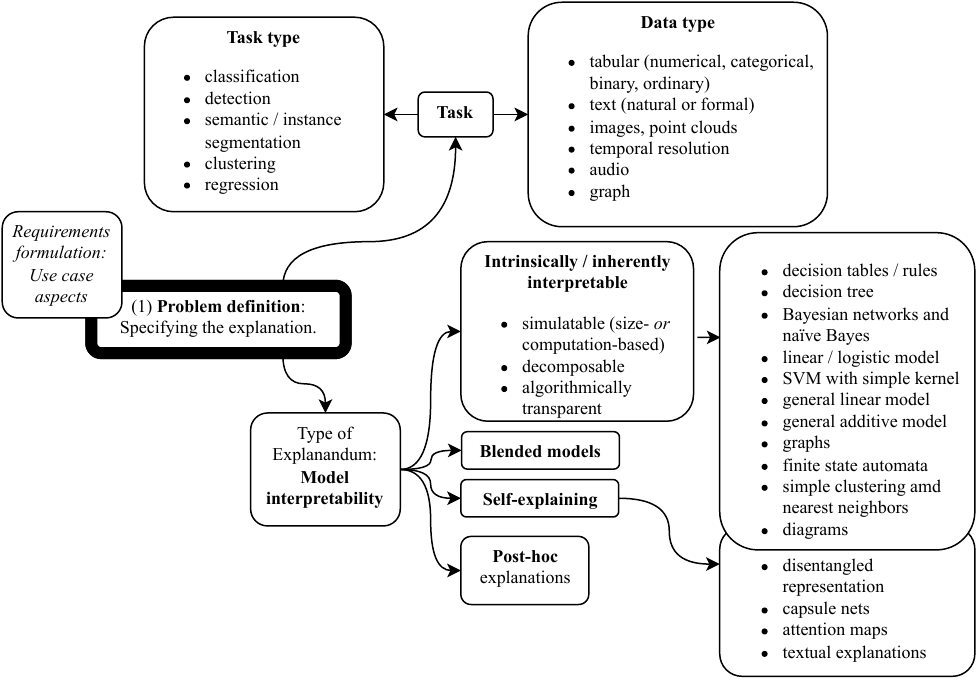}
    \caption{
        Overview of the taxonomy aspects related to the problem definition that are detailed in \autoref{sec:taxonomy.problem}.
        Find a visualization of the complete taxonomy in \autoref{fig:taxonomy} on page~\pageref{fig:taxonomy}.
        }
    \label{fig:taxonomy.problem}
\end{figure}


\subsubsection{Task}\label{sec:taxonomy.problem.task}
Out-of-the-box, XAI methods usually only apply to a specific set of
\begin{itemize}
    \item \emph{\hyperref[sec:taxonomy.problem.task.type]{task types}} of the to-be-explained model, and 
    \item \emph{\hyperref[sec:taxonomy.problem.task.data]{input data types}}. 
\end{itemize}
For white-box methods that access model internals, additional constraints
may hold for the \emph{architecture} of the model
(\cf portability aspect in \cite{yao_knowledge_2005}).

\paragraph{Task type}\label{sec:taxonomy.problem.task.type}
Typical task categories are
unsupervised clustering (clu),
regression,
classification (cls),
detection (det),
segmentation (seg), either semantic, which is pixel-wise classification,
or segmentation of instances.
Many XAI methods that target a question for classification,
\forexample, \enquote{Why this class?}
can be extended to det, seg, and temporal resolution.
This can be achieved by snippeting of the new dimensions:
\enquote{Why this class in this spatial/temporal snippet?}.
It must be noted that XAI methods working on classifiers often require
access to the prediction of a continuous classification score instead of the
final discrete classification. Such methods can also be used on regression
tasks to answer questions about local trends, \idest,
\enquote{Why does the prediction tend in this direction?}.
Examples of regression predictions are bounding box dimensions in object detection.
\begin{examples}
    \examplemethod{RISE~\cite{petsiuk_rise_2018}}
    RISE (Randomized Input Sampling for Explanation) by \longcite{petsiuk_rise_2018}
    is a model-agnostic attribution analysis method specializing in image classification.
    For an input image, it produces a heatmap that highlights those superpixels in the image,
    which, when deleted, have the greatest influence on the class confidence.
    High-attribution superpixels are found by randomly dimming superpixels of the input image.
    %
    \examplemethod{D-RISE~\cite{petsiuk_blackbox_2021}}
    The method D-RISE by \longcite{petsiuk_blackbox_2021} extends this to object detection.
    It considers not a one-dimensional class confidence but the total prediction vector of a detection. The influence of dimming is measured as the distance between the prediction vectors.
    %
    \examplemethod{ILP~\cite{cropper_turning_2020}}
    The mentioned image-specific (\idest, local) explanation methods use the
    continuous class or prediction scores of the explanandum,
    and, hence, are in principle also applicable to regressors.
    In contrast, surrogate models produced using inductive logic programming (ILP) \cite{cropper_turning_2020} require the binary classification output of a model.
    ILP frameworks require input background knowledge (logical theory),
    together with positive and negative examples. From this, a logic program in the form
    of first-order rules is learned that covers as many of the samples as possible.
    %
    \examplemethod{CA-ILP~\cite{rabold_expressive_2020}}
    An example of an ILP-based XAI method for convolutional image classifier is CA-ILP (Concept Analysis for ILP) by \longcite{rabold_expressive_2020}. In order to explain parts of the classifier with logical rules,
    they first train small global models that extract symbolic features from the DNN intermediate outputs.
    These feature outputs are then used to train an ILP surrogate model.
    %
    Lastly, clustering tasks can often be explained by providing examples or prototypes
    of the final clusters, which will be discussed in \autoref{sec:taxonomy.explanator}.
\end{examples}

\paragraph{Input data type}\label{sec:taxonomy.problem.task.data}
Not every XAI method supports every input and output \emph{signal type},
also called data type~\cite{guidotti_survey_2018}. One input type is
tabular (symbolic) data, which encompasses numerical, categorical, binary, and ordinary (ordered) data.
Other symbolic input types are natural language or graphs, and
non-symbolic types are images and point clouds (with or without temporal resolution),
as well as audio.
\begin{examples}
    Typical examples for image explanations are methods producing heatmaps.
    These highlight parts of the image that were relevant to the decision or a part thereof.
    This highlighting of input snippets can also be applied to textual inputs
    where single words or sentence parts may serve as snippets.
    \examplemethod{LIME~\cite{ribeiro_why_2016}}
    A prominent example of heatmapping that is both applicable to images and text inputs
    is the model-agnostic LIME~\cite{ribeiro_why_2016} method
    (Local Interpretable Model-agnostic Explanations).
    It locally approximates the explanandum model by a linear model on feature snippets of
    the input. For training of that linear model, randomly selected snippets are removed.
    In the case of textual inputs, the words are considered as snippets, and for images pixels are grouped into superpixels.
    The removal of superpixels is here realized by coloring them with a neutral color, e.g., black.
    %
    While LIME is suitable for image or textual input data,
    \longcite{guidotti_survey_2018} provide a broad overview of model-agnostic XAI methods
    for tabular data.
\end{examples}


\subsubsection{Model interpretability}\label{sec:taxonomy.problem.interpretability}
Model interpretability here refers to the level of interpretability of the explanandum,
\idest, the model used to solve the original task of the system.
A model is interpretable if it gives rise not only to mechanistic understanding (transparency) but also to a functional understanding by a human~\cite{paez2019pragmatic}.
Explainability of (aspects of) the explanation system can be achieved by one of the following choices:
\begin{itemize}
    \item start from the beginning with an \emph{\hyperref[sec:taxonomy.problem.interpretability.intrinsic]{intrinsically interpretable}} explanandum model
    (also called \emph{ante-hoc interpretable}~\cite{burkart_survey_2021} or \emph{intrinsically interpretable}) or a
    \item \emph{\hyperref[sec:taxonomy.problem.interpretability.blended]{blended}}, \idest, partly interpretable model
    (also called \emph{interpretable by design} \cite{burkart_survey_2021});
    \item design the explanandum model to include  \emph{\hyperref[sec:taxonomy.problem.interpretability.self]{self-explanations}} as additional output; or
    \item \emph{\hyperref[sec:taxonomy.problem.interpretability.posthoc]{post-hoc}}
    find an interpretable helper model without changing the trained explanandum model.
\end{itemize}

\paragraph{Intrinsic or inherent interpretability}\label{sec:taxonomy.problem.interpretability.intrinsic}
As introduced by \longcite{lipton_mythos_2018},
one can further differentiate between different levels of model transparency.
The model can be understood as a whole, \idest, a human can adopt it as a mental model 
(\emph{simulatable}~\cite{arrieta_explainable_2020}).
Alternatively, it can be split up into simulatable parts
(\emph{decomposable}~\cite{arrieta_explainable_2020}).
Simulatability can either be measured based on the size of the model
or based on the needed length of computation (\cf discussion of metrics in \autoref{sec:taxonomy.metrics}).
As a third category, \emph{algorithmic transparency} is considered,
which means the model is mathematically understood,
\forexample, the shape of the error surface is known.
This is considered the weakest form of transparency, because the algorithm may not be simulatable as a mental model.
%
The following models are considered inherently transparent in the literature
(\cf \cite[Chap.~4]{molnar_interpretable_2020}, \cite[Sec.~5]{guidotti_survey_2018},
\cite{nori_interpretml_2019}):
\begin{subparagraphs}
    \item[Decision tables and rules] as experimentally evaluated by \longcite{huysmans_empirical_2011,allahyari_useroriented_2011,freitas_comprehensible_2014};
        This encompasses boolean rules as can be extracted from decision trees
        or fuzzy or first-order logic rules.
        For further insights into inductive logic programming approaches to find the latter kind
        of rules, see, \forexample, the recent survey by \longcite{cropper_turning_2020}.
    \item[Decision trees] as empirically evaluated by \longcite{huysmans_empirical_2011,freitas_comprehensible_2014,allahyari_useroriented_2011};
    \item[Bayesian networks and naïve Bayes models] as of \longcite{burkart_survey_2021};
        interpretability of Bayesian network classifiers was, \forexample, experimentally evaluated by \longcite{freitas_comprehensible_2014}.
    \item[Linear and logistic models] as of, \forexample, \longcite{molnar_interpretable_2020};
    \item[Support vector machines]  as of \longcite{singh_explainable_2020};
        as long as the used kernel function is not too complex, non-linear SVMs give interesting insights into the decision boundary.
    \item[General linear models (GLM)] to inherently provide weights for the importance of input features;
        Here, it is assumed that there is a transformation, such that there is a linear relationship between the transformed input features and
        the expected output value.
        For example, in logistic regression, the transformation is the logit.
        See, \forexample, \cite[Sec.~4.3]{molnar_interpretable_2020} for a basic introduction
        and further references.
    \item[General additive models (GAM)] also inherently come with feature importance weights;
        Here, it is assumed that the expected output value is the sum of transformed features.
        See the survey by \longcite{chang_how_2020} for more details and further references.
        \begin{examples}
            \examplemethod{Additive Model Explainer~\cite{chen_explaining_2019}}
            One concrete example of general additive models is the
            Additive Model Explainer by \longcite{chen_explaining_2019}.
            They train predictors for a given set of features, and another small DNN
            predicts the additive weights for the feature predictors.
            They use this setup to learn a GAM surrogate models for a DNN,
            which also provides a prior to the weights: They should correspond to the
            sensitivity of the DNN with respect to the features.
        \end{examples}
    \item[Graphs] as of, \forexample, \longcite{xie_explainable_2020};
    \item[Finite state automata] as of \longcite{wang_empirical_2018};
    \item[Simple clustering and nearest neighbors approaches] as of \longcite{burkart_survey_2021};
        \begin{examples}
            Examples are $k$-nearest neighbors (supervised) or $k$-means clustering (unsupervised).
            \examplemethod{$k$-means clustering~\cite{hartigan_algorithm_1979}}
            The standard $k$-means clustering method introduced by \longcite{hartigan_algorithm_1979}
            works with an intuitive model, simply consisting of $k$ prototypes and
            a proximity measure, with inference associating new samples to
            the closest prototype representing a cluster.
            \examplemethod{$k$-NN~\cite{altman_introduction_1992}}
            $k$-nearest neighbors ($k$-NN) determines for a new input the $k$
            samples from a labeled database that are most similar to the new input sample.
            The majority vote of the nearest labels is then used to assign a label
            to the new instance.
            As long as the proximity measure is not too complex, these methods
            can be regarded as unsupervised respectively supervised inherently interpretable models. $k$-NN was experimentally evaluated for interpretability by \longcite{freitas_comprehensible_2014}.
        \end{examples}
    \item[Diagrams] as of \longcite{heuillet_explainability_2021}.
\end{subparagraphs}

\paragraph{Blended models}\label{sec:taxonomy.problem.interpretability.blended}
Blended models (also called \emph{interpretable by design}~\cite{burkart_survey_2021})
consist partly of intrinsically transparent, symbolic models
that are integrated in sub-symbolic non-transparent ones.
These kinds of hybrid models are especially interesting for neuro-symbolic
computing and similar fields combining symbolic with sub-symbolic models
\cite{calegari_integration_2020}.
\begin{examples}
    \examplemethod{Logic Tensor Nets~\cite{donadello_logic_2017}}
    An example of a blended model is the Logic Tensor Network.
    Their idea is to use fuzzy logic to encode logical constraints
    on DNN outputs, with a DNN acting as a fuzzy logic predicate.
    The framework by \longcite{donadello_logic_2017} allows additionally to learn
    semantic relations subject to symbolic fuzzy logic constraints.
    The relations are represented by simple linear models.
    %
    \examplemethod{%
    FoldingNet~\cite{yang_foldingnet_2017},
    Neuralized clustering~\cite{kauffmann_clustering_2019}}
    Unsupervised deep learning can be made interpretable by several approaches, \forexample,
    combining autoencoders with visualization approaches. Another approach explains
    choices of \enquote{neuralized} clustering methods \cite{kauffmann_clustering_2019}
    (\idest, clustering models translated to a DNN) with saliency maps.
    Enhancing an autoencoder was applied, for example, in the FoldingNet~\cite{yang_foldingnet_2017} architecture on point clouds.
    There, a folding-based decoder allows for viewing the reconstruction
    of point clouds, namely the warping from a 2D grid into the point cloud surface.
    A saliency-based solution can be produced by algorithms such as layer-wise relevance propagation, which will be discussed in later examples.
\end{examples}

\paragraph{Self-explaining models}\label{sec:taxonomy.problem.interpretability.self}
Self-explaining models provide additional outputs that explain the output
of a single prediction.
According to \longcite{gilpin_explaining_2018}, there are three standard types of
outputs of explanation generating models:
\emph{attention maps},
\emph{disentangled representations}, and
\emph{textual or multi-modal explanations}.
\begin{subparagraphs}[font=\appenddot]
    \item[Attention maps]
    These are heatmaps that highlight the relevant parts of a given single input
    for the respective output.
    \begin{examples}
    The work by \longcite{kim_interpretable_2017} adds an attention module to a DNN
    that is processed in parallel to, and later multiplied with, convolutional outputs.
    Furthermore, they suggest a clustering-based post-processing of the
    attention maps to highlight the most meaningful parts.    
    \end{examples}
    
    \item[Disentangled representations]
    Representations in the intermediate output of the explanandum are called disentangled
    if single or groups of dimensions therein directly represent symbolic (also called semantic) concepts.
    \begin{examples}
        One can, by design, force one layer of a DNN to exhibit a disentangled representation.
        \examplemethod{Capsule Nets~\cite{sabour_dynamic_2017}}
        One example is the capsule network by \longcite{sabour_dynamic_2017},
       where groups of neurons,
        the capsules, characterize each an individual entity, \forexample, an object
        or object part.
        The length of a capsule vector is interpreted as the probability that
        the corresponding object is present, while the rotation encodes the properties
        of the object (\forexample, rotation or color).
        Later capsules get as input the weighted sum of transformed previous capsule outputs,
        with the transformations learned and the weights obtained in an iterative
        routing process.
        A simpler disentanglement than an alignment of semantic concepts with groups
        of neurons is the alignment of single dimensions.
        \examplemethod{ReNN~\cite{wang_renn_2018}}
        This is done, \forexample, in the ReNN architecture developed by \longcite{wang_renn_2018}.
        They explicitly modularize their DNN to ensure semantically meaningful
        intermediate outputs.
        \examplemethod{Semantic Bottlenecks~\cite{losch_interpretability_2019}}
        Other methods rather follow a post-hoc approach that fine-tunes a trained
        DNN towards more disentangled representations, as suggested for
        Semantic Bottleneck Networks~\cite{losch_interpretability_2019}.
        These consist of the pretrained backbone of a DNN, proceeded by a layer
        in which each dimension corresponds to a semantic concept, called semantic bottleneck,
        and finalized by a newly trained front DNN part. During fine-tuning,
        first, the connections from the backend to the semantic bottleneck are trained,
        then the parameters of the front DNN.
        \examplemethod{Concept Whitening~\cite{chen_concept_2020}}
        Another interesting fine-tuning approach is that of
        concept whitening by \longcite{chen_concept_2020}, which supplements
        batch-normalization layers with a linear transformation that learns to align
        semantic concepts with unit vectors of an activation space.
    \end{examples}
    
    \item[Textual or multi-model explanations]
    These provide the explainee with a direct verbal or combined explanation as part of the model output.
    \begin{examples}
        \examplemethod{\cite{kim_textual_2018}}
        An example are the explanations provided by \longcite{kim_textual_2018}
        for the application of end-to-end steering control in autonomous driving.
        Their approach is two-fold: They add a custom layer that produces
        attention heatmaps similar to those from \cite{kim_interpretable_2017};
        a second custom part uses these heatmaps to generate
        textual explanations of the decision, which are (weakly)
        aligned with the model processing.
        \examplemethod{ProtoPNet~\cite{chen_this_2019}}
        ProtoPNet by \longcite{chen_this_2019} for image classification provides visual
        examples rather than text. The network architecture is based on first
        selecting prototypical image patches and then inserting a prototype layer
        that predicts similarity scores for patches of an instance with prototypes.
        These can then be used for explanation of the final result in the manner of
        \enquote{This is a sparrow as its beak looks like that of other sparrow examples}.
        %
        \examplemethod{\cite{hendricks_generating_2016}}
        A truly multi-modal example is that by \longcite{hendricks_generating_2016}, which
        trains alongside a classifier a long-short term memory DNN (LSTM) to generate
        natural language justifications of the classification.
        The LSTM uses both the intermediate features and predictions of the image classifier
        and is trained towards high-class discriminativeness of the justifications.
        The explanations can optionally encompass bounding boxes for features
        that were important for the classification decision, making it multi-modal.
    \end{examples}
\end{subparagraphs}

\paragraph{Post-hoc}\label{sec:taxonomy.problem.interpretability.posthoc}
Post-hoc methods use a (local or global) helper model from which to derive an explanation.
This explainable helper model can either aim to
\begin{itemize}
    \item fully mimic the behavior of the explanandum, or
    \item only approximate sub-aspects like input attribution.
\end{itemize}
Helper models that fully approximate the explanandum are often called \emph{surrogate} or \emph{proxy} models,
and the process of training them is termed \emph{model distillation}, \emph{student-teacher} approach, or \emph{model induction}.
However, it is often hard to differentiate between the two types.
Hence, for consistency, we here use the terms proxy and surrogate model for any type of helper model.
Many examples of post-hoc methods are given in the course of the upcoming taxonomy aspects.


\subsection{Explanator} \label{sec:taxonomy.explanator}
\begin{figure}
    \centering
    \includegraphics[scale=\thetaxonomygraphicscale]{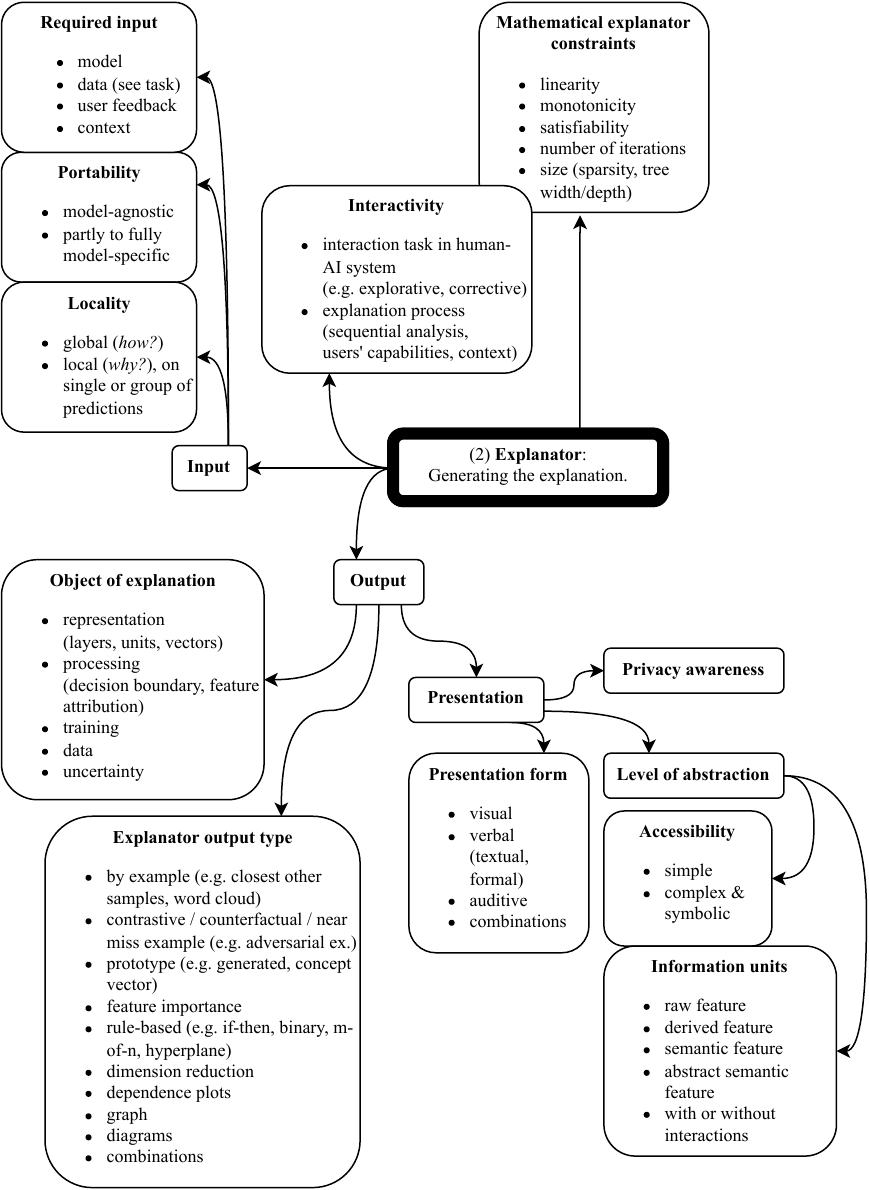}
    \caption{
        Overview of the taxonomy aspects related to the explanator that are detailed in \autoref{sec:taxonomy.metrics}.
        Find a visualization of the complete taxonomy in \autoref{fig:taxonomy} on page~\pageref{fig:taxonomy}.
        }
    \label{fig:taxonomy.explanator}
\end{figure}

One can consider an explanator simply as an implemented function that outputs explanations.
This allows to structure aspects of the explanator into the main defining aspects of a function:
\begin{itemize}
    \item the \emph{\hyperref[sec:taxonomy.explanator.input]{input}},
    \item the \emph{\hyperref[sec:taxonomy.explanator.output]{output}},
    \item the function class described by \emph{\hyperref[sec:taxonomy.explanator]{mathematical properties or constraints}}, and
    \item the actual processing of the explanation function,
    like its \emph{\hyperref[sec:taxonomy.explanator.interactivity]{interactivity}}.
\end{itemize}
An overview of the aspects discussed in this section is given in \autoref{fig:taxonomy.explanator}.

\subsubsection{Input}\label{sec:taxonomy.explanator.input}
The following explanator characteristics are related to the explanator input:
\begin{itemize}
    \item What are the \emph{\hyperref[sec:taxonomy.explanator.input.required]{required inputs}} (the explanandum model, data samples, or even user feedback)?
    \item In how far is the method \emph{\hyperref[sec:taxonomy.explanator.input.portability]{portable}} to other input types, \forexample, explanandum model types?
    \item In how far are explanations \emph{\hyperref[sec:taxonomy.explanator.input.locality]{local to an input instance or global for the complete input}}?
\end{itemize}

\paragraph{Required input}\label{sec:taxonomy.explanator.input.required}
The necessary inputs to the explanator may differ amongst methods \cite{spinner_explainer_2020}.
The explanandum must usually be provided to the explanator.
Many methods do also require valid \emph{data} samples. Some even require
\emph{user feedback} (\cf \autoref{sec:taxonomy.explanator.interactivity}) or
further situational \emph{context} (\cf \cite{dey_understanding_2001} for a more detailed definition of context).

\paragraph{Portability}\label{sec:taxonomy.explanator.input.portability}
An important practical aspect of post-hoc explanations is whether or how far
the explanation method is dependent on access to the internals of the explanandum model.
This level of dependency is called portability, translucency, or transferability.
In the following, we will not further differentiate between the strictness of
requirements of model-specific methods.
Transparent and self-explaining models are always model-specific, as
the interpretability requires a special model type or model architecture (modification).
Higher levels of dependency are:
\begin{subparagraphs}
    \item[Model-agnostic]
    also called \emph{pedagogical}~\cite{yao_knowledge_2005} or black-box:
    This means that only access to model input and output is required.
    \begin{examples}
        A prominent example of model-agnostic methods is the previously discussed
        LIME~\cite{ribeiro_why_2016} method for local approximation via a linear model.
        \examplemethod{SHAP~\cite{lundberg_unified_2017}}
        Another method to find feature importance weights without any access
        to model internals is SHAP (SHapley Additive exPlanation) by \longcite{lundberg_unified_2017}.
        Their idea is to axiomatically ensure: local fidelity; features missing
        from the original input have no effect; an increase in weight
        also means an increased attribution of the feature to the final output
        and uniqueness of the weights. Just as LIME, SHAP just requires a
        definition of \enquote{feature} or snippet on the input in order to
        be applicable.
    \end{examples}
    
    \item[Model-specific]
    also called \emph{decompositional}~\cite{yao_knowledge_2005} or white-box:
    This means that access is needed to the internal processing or architecture of
    the explanandum model, or even constraints apply.
    \begin{examples}
        Methods relying on gradient or relevance information for the generation of
        visual attention maps are strictly model-specific.
        \examplemethod{Sensitivity Analysis~\cite{baehrens_how_2010}}
        A gradient-based method is Sensitivity Analysis by \longcite{baehrens_how_2010}.
        They pick the vector representing the steepest ascension in the gradient
        tangential plane of a sample point. This method is independent of the
        type of input features but can only analyze a single
        one-dimensional output at once.
        %
        \examplemethod{%
        Deconvnet~\cite{zeiler_visualizing_2014},
        Backprop~\cite{simonyan_deep_2014},
        Guided Backprop~\cite{springenberg_striving_2015}}
        Deconvnet by \longcite{zeiler_visualizing_2014} instead is agnostic to the type of output, but depends on a convolutional architecture and image inputs.
        The same holds for its successors
        Backpropagation~\cite{simonyan_deep_2014} and Guided Backpropagation~\cite{springenberg_striving_2015}.
        They approximate a reconstruction of input by defining inverses of pool
        and convolution operations. This allows for backpropagating the activation of 
        single filters back to input image pixels
        (see \cite{weitz_applying_2018} for a good overview).
        \examplemethod{LRP~\cite{bach_pixelwise_2015}}
        The idea of Backpropagation is generalized axiomatically by LRP
        (Layer-wise Relevance Propagation): \longcite{bach_pixelwise_2015} require that the sum of linear relevance weights
        for each neuron in a layer should be constant throughout the layers. The rationale behind this is that relevance is neither created nor extinguished from layer to layer.
        Methods that achieve this are, \forexample, Taylor decomposition or
        the backpropagation of relevance weighted by the forward-pass weights.
        \examplemethod{PatternAttribution~\cite{kindermans_learning_2018}}
        The advancement PatternAttribution by \longcite{kindermans_learning_2018} fulfills
        the additional constraint to be sound on linear models.
    \end{examples}
    
    \item[Hybrid]
    also called \emph{eclectic}~\cite{yao_knowledge_2005} or \emph{gray-box}:
    This means the explanator only depends on access to parts of the model
    intermediate output, but not the full architecture.
    \begin{examples}
        \examplemethod{DeepRED~\cite{zilke_deepred_2016}}
        The rule extraction technique DeepRED
        (Deep Rule Extraction with Decision tree induction) by \longcite{zilke_deepred_2016} is an example of
        an eclectic method, so neither fully model-agnostic nor totally
        reliant on access to model internals. The approach conducts a backward
        induction over the layer outputs of a DNN, between each two applying
        a decision tree extraction. While they enable rule extraction for arbitrarily
        deep DNNs, only small networks will result in rules of decent length
        for explanations.
    \end{examples}
\end{subparagraphs}

\paragraph{Explanation locality}\label{sec:taxonomy.explanator.input.locality}
Literature differentiates between different ranges of validity of an explanation
respectively surrogate model. A surrogate model is valid in the ranges where high
fidelity can be expected (see \autoref{sec:taxonomy.metrics}).
The range of input required by the explanator depends on the targeted validity range,
so whether the input must represent a \emph{local} or the \emph{global} behavior
of the explanandum.
The general locality types are:
\begin{subparagraphs}[font=\appenddot]
    \item[Local]
    An explanation is considered local if the explanator is valid in a neighborhood of one or
    a group of given (valid) input samples.
    Local explanations tackle the question of \emph{why} a given decision
    for one or a group of examples was made.
    \begin{examples}
        Heatmapping methods are typical examples for local-only explanators, such as
        the discussed perturbation-based model-agnostic methods
        RISE~\cite{petsiuk_rise_2018},
        D-RISE~\cite{petsiuk_blackbox_2021},
        LIME~\cite{ribeiro_why_2016},
        SHAP~\cite{lundberg_unified_2017},
        as well as the model-specific sensitivity and backpropagation based methods
        LRP~\cite{bach_pixelwise_2015},
        PatternAttribution~\cite{kindermans_learning_2018},
        Sensitivity Analysis~\cite{baehrens_how_2010},
        and Deconvnet and its successors~\cite{zeiler_visualizing_2014,simonyan_deep_2014,springenberg_striving_2015}.
    \end{examples}
    
    \item[Global]
    An explanation is considered global if the explanator is valid in the complete (valid) input space.
    Other than the \emph{why} of local explanations, global interpretability can
    also be described as answering \emph{how} a decision is made.
    \begin{examples}
        \examplemethod{Explanatory Graphs~\cite{zhang_interpreting_2018}}
        A graph-based global explanator is generated by \longcite{zhang_interpreting_2018}.
        Their idea is that semantic concepts in an image usually
        consist of sub-objects to which they have a constant relative spatial relation
        (\forexample, a face has a nose in the middle and two eyes next to each other) and that
        the localization of concepts should not only rely on high filter activation patterns,
        but also on their sub-part arrangement.
        To achieve this, they translate the convolutional layers of a DNN into a tree of nodes (concepts),
        the \emph{explanatory graph}.
        Each node belongs to one filter, is anchored at a fixed spatial position in the image,
        and represents a spatial arrangement of its child nodes.
        The graph can also be used for local explanations via heatmaps.
        For localizing a node in an input image, the node is assigned the position closest
        to its anchor for which (1)~its filter activation is highest, and (2)~the
        expected spatial relation to its children is best fulfilled.
        \examplemethod{Feature Visualization~\cite{olah_feature_2017}}
        While most visualization-based methods provide only local visualizations,
        Feature Visualizations as reviewed by \longcite{olah_feature_2017} give a global, prototype-based, visual explanation.
        The goal here is to visualize the functionality of a DNN's part.
        It is achieved by finding prototypical input examples that strongly activate that part.
        These can be found via picking, search, or optimization.
        %
        \examplemethod{VIA~\cite{thrun_extracting_1995}}
        Other than visualizations, rule extraction methods usually only provide
        global approximations.
        An example is the well-known model-agnostic rule extractor VIA
        (Validity Interval Analysis) by \longcite{thrun_extracting_1995}, which iteratively refines or generalizes pairs of
        input- and output-intervals.
        %
        \examplemethod{SpRAy~\cite{lapuschkin_unmasking_2019}}
        An example of getting from local to global explanations is
        SpRAy (Spectral Relevance Analysis) by \longcite{lapuschkin_unmasking_2019}.
        They suggest to apply spectral clustering~\cite{vonluxburg_tutorial_2007}
        to local feature attribution heatmaps of data samples in order to find
        spuriously distinct global behavioral patterns.
        The heatmaps were generated via LRP~\cite{bach_pixelwise_2015}.
    \end{examples}
\end{subparagraphs}


\subsubsection{Output}\label{sec:taxonomy.explanator.output}
The output is characterized by several aspects:
\begin{itemize}
    \item what is explained (the \emph{\hyperref[sec:taxonomy.explanator.output.object]{object of explanation}}),
    \item how it is explained (the actual \emph{\hyperref[sec:taxonomy.explanator.output.type]{output type}}, also called \emph{explanation content type}), and
    \item how it is \emph{\hyperref[sec:taxonomy.explanator.output.presentation]{presented}}.
\end{itemize}

\paragraph{Object of explanation}\label{sec:taxonomy.explanator.output.object}
The object (or scope~\cite{molnar_interpretable_2020}) of an explanation describes
which item of the development process should be explained.
Items we identified in the literature:
\begin{subparagraphs}[font=\appenddot]
    \item[Processing]
    The objective is to understand the (symbolic) processing pipeline of the model,
    \idest, to answer parts of the question \enquote{How does the model work?}.
    This is the usual case for model-agnostic analysis methods.
    Types of processing to describe are, \forexample,
    the \emph{decision boundary} and \emph{feature attribution} (or feature importance).
    Note that these are closely related, as highly important features usually
    locally point out the direction to the decision boundary.
    In case a symbolic explanator is targeted, one may need to first find a
    symbolic representation of the input, output, or the model's internal representation.
    Note that model-agnostic methods that do not investigate the
    input data usually target explanations of the model processing.
    \begin{examples}
        Feature attribution methods encompass all the discussed
        attribution heatmapping methods (\forexample, 
        RISE~\cite{petsiuk_rise_2018},
        LIME~\cite{ribeiro_why_2016},
        LRP~\cite{bach_pixelwise_2015}).
        %
        LIME can be considered a corner case: In addition to explaining feature
        importance it approximates the decision boundary using a
        linear model on superpixels. The linear model itself may already serve as an
        explanation.
        Typical ways to describe decision boundaries are decision trees
        or sets of rules, as extracted by the discussed
        VIA~\cite{thrun_extracting_1995}, and
        DeepRED~\cite{zilke_deepred_2016} approaches.
        \examplemethod{TREPAN~\cite{craven_extracting_1995}}
        Standard candidates for model-agnostic decision tree extraction are
        TREPAN by \longcite{craven_extracting_1995}, and C4.5 by \longcite{quinlan_c4_1993}.
        TREPAN uses M-of-N rules at the split points of the extracted decision tree.
        \examplemethod{C4.5~\cite{quinlan_c4_1993}}
        C4.5 uses interval-based splitting points, and generates shallower but wider trees compared to TREPAN.
        \examplemethod{Concept Tree~\cite{renard_concept_2019}}
        Concept tree is a recent extension of TREPAN by \longcite{renard_concept_2019} that adds
        automatic grouping of correlated features into the candidate concepts to use for
        the tree nodes.
    \end{examples}
    
    \item[Inner representation]
    Machine learning models learn new representations of the input space, like the
    latent space representations found by DNNs.
    Explaining these inner representations answers, \enquote{How does the model see the world?}.
    A more fine-grained differentiation considers whether
    \emph{layers}, \emph{units}, or \emph{vectors} in the feature space 
    are explained.
    \begin{examples}
        \begin{itemize}
            \item Units:
            One example of unit analysis is the discussed Feature Visualization~\cite{olah_feature_2017}.
            \examplemethod{NetDissect~\cite{bau_network_2017}}
            In contrast to this unsupervised assignment of convolutional filters to
            prototypes, NetDissect (Network Dissection) by \longcite{bau_network_2017}
            assigns filters to pre-defined semantic concepts in a supervised manner:
            For a filter, that semantic concept (color, texture, material, object, or object part)
            is selected for which the ground truth segmentation masks have the highest overlap
            with the upsampled filter's activations. The authors also suggest that concepts that
            are less entangled, so less distributed over filters, are more interpretable,
            which is measurable with their filter-to-concept-alignment technique.
            
            \item Vectors:
            \examplemethod{Net2Vec~\cite{fong_net2vec_2018}}
            Other than NetDissect, Net2Vec by \longcite{fong_net2vec_2018} also wants to assign concepts
            to their possibly entangled representations in the latent space.
            Given a concept, they train a linear $1\times1$-convolution on the output of a layer to segments the respective concept with an image.
            The weight vector of the linear model for a concept can be understood as
            a prototypical representation (embedding) for that concept in the DNN
            intermediate output. They found that such embeddings behave like vectors in
            a word vector space: Concepts that are semantically similar feature embeddings
            with high cosine similarity.
            \examplemethod{TCAV~\cite{kim_interpretability_2018}}
            Similar to Net2Vec, TCAV
            (Testing Concept Activation Vectors) also aims to find embeddings of NetDissect concepts.
            \longcite{kim_interpretability_2018} are not interested in embeddings that are represented as a linear combination of
            convolutional filters, but instead in embedding vectors lying in the space of
            the complete layer output. In other words, they do not segment concepts,
            but make an image-level classification of whether the concept is present.
            These are found by using an SVM model instead of the $1\times1$-convolution.
            Additionally, they suggest using partial derivatives along those concept vectors
            to find the local attribution of a semantic concept to a certain output.
            \examplemethod{ACE~\cite{ghorbani_automatic_2019}}
            Other than the already mentioned supervised methods,
            ACE (Automatic Concept-based Explanations) by by \longcite{ghorbani_automatic_2019}
            does not learn a linear classifier but does an unsupervised clustering
            of concept candidates in the latent space.
            The cluster center is then selected as the embedding vector.
            A superpixeling approach is used together with outlier removal to obtain
            concept candidates.
            
            \item Layers:
            \examplemethod{Concept completeness~\cite{yeh_completenessaware_2020}, IIN~\cite{esser_disentangling_2020}}
            The works of \longcite{yeh_completenessaware_2020} and
            the IIN (invertible interpretation networks) approach by \longcite{esser_disentangling_2020}
            extend on the previous approaches and analyze a complete layer output
            space at once. For this, they find a subspace with a basis of concept embeddings,
            which allows an invertible transformation to a disentangled representation space.
            While IIN uses invertible DNNs for the bijection of concept space to latent space,
            \longcite{yeh_completenessaware_2020} use linear maps in their experiments.
            These approaches can be seen as a post-hoc version of the
            Semantic Bottleneck~\cite{losch_interpretability_2019} architecture,
            only not replacing the complete later part of the model, but just learning
            connections from the bottleneck to the succeeding trained layer.
            \longcite{yeh_completenessaware_2020} additionally introduce the notion of
            completeness of a set of concepts as the maximum performance of the model
            intercepted by the semantic bottleneck.
        \end{itemize}
    \end{examples}
    
    \item[Development (during training)]
    Some methods focus on assessing the effects during training \cite[Sec.~2.3]{molnar_interpretable_2020}:
    \enquote{How does the model evolve during the training? What effects do new samples have?}
    \begin{examples}
        \examplemethod{\cite{shwartz-ziv_opening_2017}}
        One example is the work of \longcite{shwartz-ziv_opening_2017}, who inspect the model
        during training to investigate the role of depth in neural networks.
        Their findings indicate that depth actually is of computational benefit.
        %
        \examplemethod{Influence Functions~\cite{koh_understanding_2017}}
        An example which can be used to provide, \forexample, prototypical explanations
        are Influence Functions by \longcite{koh_understanding_2017}.
        They gather the influence of training samples during the training to later
        assess the total impact of samples on the training.
        They also suggest using this information as a proxy to estimate the influence of the samples on model decisions.
    \end{examples}
    
    \item[Uncertainty] \longcite{molnar_interpretable_2020}
    suggests to capture and explain (\forexample, visualize) the uncertainty of a prediction of the model.
    This encompasses the broad field of Bayesian deep learning \cite{kendall_what_2017}
    and uncertainty estimation \cite{henne_benchmarking_2020}.
    Several works argue why it is important
    to make the uncertainty of model decisions accessible to users.
    For example, \longcite{poceviciute_survey_2020} argues this for medical applications, and \longcite{mcallister_concrete_2017} for autonomous driving.
    
    \item[Data]
    Pre-model interpretability \cite{carvalho_machine_2019} is the point where explainability
    touches the large research area of data analysis and feature mining.
    \begin{examples}
        \examplemethod{PCA~\cite{jolliffe_principal_2002}}
        Typical examples for projecting high-dimensional data into easy-to-visualize
        2D space are component analysis methods like PCA (Principal Component Analysis) which was introduced by \longcite{jolliffe_principal_2002}.
        \examplemethod{t-SNE~\cite{maaten_visualizing_2008}}
        A slightly more sophisticated approach is t-SNE (t-Distributed Stochastic Neighbor Embedding) by \longcite{maaten_visualizing_2008}.
        In order to visualize a set of high-dimensional data points,
        they try to find a map from these points into a 2D or 3D space
        that is faithful to pairwise similarities.
        %
        \examplemethod{Spectral Clustering~\cite{vonluxburg_tutorial_2007}}
        And also clustering methods can be used to generate prototype-
        or example-based explanations of typical features in the data.
        Examples here are k-means clustering~\cite{hartigan_algorithm_1979}
        and graph-based spectral clustering~\cite{vonluxburg_tutorial_2007}.
    \end{examples}
\end{subparagraphs}

\paragraph{Output type}\label{sec:taxonomy.explanator.output.type}
The output type, also considered the actual explanator~\cite{guidotti_survey_2018},
describes the type of information presented to the explainee.
Note that this (\enquote{what} is shown) is mostly independent of
the presentation form (\enquote{how} it is shown).
Typical types are:
\begin{subparagraphs}
    \item[By example instance,]
    \forexample, closest other samples, word cloud;
    \begin{examples}
        The discussed ProtoPNet~\cite{chen_this_2019} is based on selecting
        and comparing relevant example snippets from the input image data.
    \end{examples}
    
    \item[Contrastive / counterfactual / near miss examples,]
    including adversarial examples;
    The goal here is to explain for an input why the respective output was as obtained instead of a desired output. This is done by presenting how the input features have to change in order to obtain the alternative output.
    Counterfactual examples are sometimes seen as a special case of more general contrastive examples \cite{stepin_survey_2021}.
    Desirables associated specifically with counterfactual examples are that they are valid inputs close to the original examples and with few features changed (\emph{sparsity}) that are actionable for the explainee and that they adhere to known causal relations \cite{guidotti_principles_2021,verma_counterfactual_2020,keane_if_2021}.
    \begin{examples}
        \examplemethod{CEM~\cite{dhurandhar_explanations_2018}}
        The perturbation-based feature importance heatmapping approach of RISE is extended in
        CEM (Contrastive, Black-box Explanations Model) by \longcite{dhurandhar_explanations_2018}.
        They do not only find positively contributing features but also the
        features that must minimally be absent to not change the output.
    \end{examples}
    
    \item[Prototype,]
    \forexample, generated, concept vector;
    \begin{examples}
        A typical prototype generator is used in the discussed
        Feature Visualization method \cite{olah_feature_2017}:
        images are generated, \forexample, via gradient descent, that represent the
        prototypical pattern for activating a filter.
        While this considers prototypical inputs, concept embeddings as collected in
        TCAV~\cite{kim_interpretability_2018} and Net2Vec~\cite{fong_net2vec_2018}
        describe prototypical activation patterns for a given semantic concept.
        The concept mining approach ACE~\cite{ghorbani_automatic_2019} combines
        prototypes with examples: They search a concept embedding as a prototype for
        an automatically collected set of example patches, that, in turn, can be used to
        explain the prototype.
    \end{examples}
    
    \item[Feature importance] that will highlight features with high attribution or influence on the output;
    \begin{examples}
        A lot of feature importance methods producing heatmaps have been discussed
        before, such as
        RISE~\cite{petsiuk_rise_2018},
        D-RISE~\cite{petsiuk_blackbox_2021},
        CEM~\cite{dhurandhar_explanations_2018},
        LIME~\cite{ribeiro_why_2016},
        SHAP~\cite{lundberg_unified_2017},
        LRP~\cite{bach_pixelwise_2015},
        PatternAttribution~\cite{kindermans_learning_2018},
        Sensitivity Analysis~\cite{baehrens_how_2010},
        Deconvnet and successors~\cite{zeiler_visualizing_2014,simonyan_deep_2014,springenberg_striving_2015}%
        .
        \examplemethod{\cite{fong_interpretable_2017}}
        One further example is the work by \longcite{fong_interpretable_2017}, who follow a
        perturbation-based approach.
        Similar to RISE, their idea is to find a minimal occlusion mask that, if used to
        perturb the image (\forexample, blur, noise, or blacken), maximally changes the outcome.
        To find the mask, backpropagation is used, making it a model-specific method.
        %
        \examplemethod{%
        CAM~\cite{zhou_learning_2016},
        Grad-CAM~\cite{selvaraju_gradcam_2017}}
        Some older but popular and simpler example methods are
        Grad-CAM by \longcite{selvaraju_gradcam_2017} and its predecessor
        CAM (Class Activation Mapping) by \longcite{zhou_learning_2016}.
        While Deconvnet and its successors can only consider the feature importance
        with respect to intermediate outputs, (Grad-)CAM produces class-specific heatmaps,
        which are the weighted sum of the filter activation maps for one (usually the last)
        convolutional layer.
        For CAM, it is assumed the convolutional backend is finalized by a global
        average pooling layer that densely connects to the final classification output.
        Here, the weights in the sum are the weights connecting the neurons of the global average pooling layer to the class outputs. For Grad-CAM, the weights in the sum are the averaged derivation of the class output by each activation map pixel.
        %
        \examplemethod{Concept-wise Grad-CAM~\cite{zhou_interpretable_2018}}
        This is also used in the more recent work of \longcite{zhou_interpretable_2018},
        who do not apply Grad-CAM directly to the output but to each of a minimal set
        of projections from a convolutional intermediate output of a DNN that predict
        semantic concepts.
        %
        \examplemethod{SIDU~\cite{muddamsetty_introducing_2021}}
        Similar to Grad-CAM, SIDU (Similarity Distance and Uniqueness) by \longcite{muddamsetty_introducing_2021}
        also adds up the filter-wise weighted activations of the last convolutional layer.
        The weights encompass a combination of a similarity score and a uniqueness score for the prediction output under each filter activation mask.
        The scores aim for high similarity of a masked prediction with the original
        one and low similarity to the other masked prediction, leading to masks
        capturing more interesting object regions.
    \end{examples}
    
    \item[Rule-based,]
    \forexample, decision tree; or if-then, binary, m-of-n, or  hyperplane rules (\cf \cite{hailesilassie_rule_2016});
    \begin{examples}
        The mentioned exemplary rule-extraction methods
        DeepRED~\cite{zilke_deepred_2016} and
        VIA~\cite{thrun_extracting_1995}, as well as 
        decision tree extractors
        TREPAN~\cite{craven_extracting_1995},
        Concept Tree~\cite{renard_concept_2019}, and
        C4.5~\cite{quinlan_c4_1993}
        all provide global, rule-based output.
        For further rule extraction examples, we refer the reader to the comprehensive surveys
        \longcite{hailesilassie_rule_2016,wang_comparative_2018,augasta_rule_2012} on the topic
        and the survey by \longcite{wang_empirical_2018} for recurrent DNNs.
        %
        \examplemethod{LIME-Aleph~\cite{rabold_explaining_2018}}
        An example of a local rule-extractor is the recent LIME-Aleph
        approach by \longcite{rabold_explaining_2018}, which generates a local explanation in the form of first-order logic rules.
        This is learned using inductive logic programming (ILP)~\cite{cropper_turning_2020}
        trained on the symbolic knowledge about a set of semantically similar examples.
        Due to the use of ILP, the approach is limited to tabular input data and
        classification outputs, but just like LIME, it is model-agnostic.
        %
        \examplemethod{NBDT~\cite{wan_nbdt_2020}}
        A similar approach is followed by NBDT (Neural-Backed Decision Trees).
        Here, \longcite{wan_nbdt_2020} assume that the concept embeddings of super-categories are
        represented by the mean of their sub-category vectors
        (\forexample, the mean of \enquote{cat} and \enquote{dog} should be
        \enquote{animal with four legs}).
        This is used to infer from bottom-to-top a decision tree where the nodes are
        super-categories, and the leaves are the classification classes.
        At each node, it is decided which of the sub-nodes best applies to the image.
        As embedding for a leaf concept (an output class), they suggest taking
        the weights connecting the penultimate layer to a class output,
        and as similarity measure for the categories, they use dot-product
        (\cf Net2Vec and TCAV).
    \end{examples}
    
    \item[Dimension reduction,]
    \idest, sample points are projected to a sub-space;
    \begin{examples}
        Typical dimensionality reduction methods mentioned previously are
        PCA~\cite{jolliffe_principal_2002} and
        t-SNE~\cite{maaten_visualizing_2008}.
    \end{examples}
    
    \item[Dependence plots]
    which plot the effect of an input feature on the final output of a model (\cf \cite{adadi_peeking_2018,carvalho_machine_2019});
    \begin{examples}
        \examplemethod{PDP~\cite{friedman_greedy_2001}}
        PDP (Partial Dependency Plots, \cf \cite[sec.~5.1]{molnar_interpretable_2020})
        by \longcite{friedman_greedy_2001}
        calculate for one input feature, and for each value of this feature,
        the expected model outcome is averaged over the dataset.
        This results in a plot (for each output) that indicates the
        global influence of the respective feature on the model.
        %
        \examplemethod{ICE~\cite{goldstein_peeking_2015}}
        The local equivalent by \longcite{goldstein_peeking_2015}, ICE
        (Individual Conditional Expectation, \cf \cite[sec.~5.2]{molnar_interpretable_2020}) plots,
        obtain the PDP for generated data samples locally around a given sample.
    \end{examples}
    
    \item[Graphs] as of \forexample \longcite{mueller_explanation_2019,mueller_principles_2021,linardatos_explainable_2021};
    \begin{examples}
        The previously discussed Explanatory Graph~\cite{zhang_interpreting_2018}
        method provides, amongst others, a graph-based explanation output.
    \end{examples}
    
    \item[Combinations] of mentioned model types.
    
\end{subparagraphs}

\paragraph{Presentation}\label{sec:taxonomy.explanator.output.presentation}
The presentation of information can be characterized by two categories of
properties:
the used \emph{presentation form} and
the \emph{level of abstraction} used to present available information.
The presentation form simply summarizes the human sensory input channels
utilized by the explanation, which can be:
visual (the most common one including diagrams, graphs, and heatmaps),
textual in either natural language or formal form,
auditive, and
combinations thereof.
In the following, the aspects influencing the level of abstraction are elaborated.
These can be split up into
(1) aspects of the smallest building blocks of the explanation, the \emph{information units},
and
(2) the \emph{accessibility} or level of \emph{complexity} of their combinations (the information units).
Lastly, further filtering may be applied before finally presenting the explanation,
including privacy filters. 
\begin{subparagraphs}[font=\appenddot]
    \item[Information units]
    The basic units of the explanation, cognitive chunks \cite{doshi-velez_rigorous_2017}, or information units,
    may differ in the level of processing applied to them.
    The simplest form are unprocessed \emph{raw features}, as used
    in explanations by example.
    \emph{Derived features} capture some indirect information contained in
    the raw inputs, like superpixels or attention heatmaps.
    These need not necessarily have semantic meaning to the explainee,
    in contrast to explicitly \emph{semantic features}, 
    \forexample, concept activation vector attributions.
    The last type of information unit are \emph{abstract semantic features}
    not directly grounded in any input, \forexample, generated prototypes.
    \emph{Feature interactions} may occur as information units or be left unconsidered
    for the explanation.
    \begin{examples}
        Some further notable examples of heatmapping methods for feature attribution
        are SmoothGrad by \longcite{smilkov_smoothgrad_2017} and 
        Integrated Gradients by \longcite{sundararajan_axiomatic_2017}.
        One drawback of the methods described so far is that they linearly approximate the loss surface in a point-wise manner.
        Hence, they struggle with \enquote{rough} loss surfaces that
        exhibit significant variation in the point-wise values, gradients, and thus feature importance \cite{samek_interpretable_2020}.
        \examplemethod{SmoothGrad~\cite{smilkov_smoothgrad_2017}}
        SmoothGrad
        aims to mitigate this by averaging the gradient from random samples
        within a ball around the sample to investigate.
        \examplemethod{Integrated Gradients~\cite{sundararajan_axiomatic_2017}}
        Integrated gradients do the averaging
        (to be precise: integration) along a path between two points in the input space.
        %
        \examplemethod{Integrated Hessians~\cite{janizek_explaining_2020}}
        A technically similar approach but with a different goal is
        Integrated Hessians~\cite{janizek_explaining_2020}.
        They intend not to grasp and visualize the sensitivity of the model for
        one feature (as a derived feature), but their information units are interactions
        of features, \idest, how much the change of one feature changes the
        influence of the other on the output.
        This is done by having a look at the Hessian matrix, which is obtained
        by two subsequent Integrated Gradients calculations.
    \end{examples}
    
    \item[Accessibility]
    The accessibility, level of detail, or level of complexity describes
    how much intellectual effort the explainee has to bring up in order to
    understand the simulatable parts of the explanation.
    Thus, the perception of complexity heavily depends on the end-user, which is mirrored in the human-grounded complexity / interpretability metric discussed later in \autoref{sec:taxonomy.metrics}.
    In general, one can differentiate between representations
    that are considered \emph{simpler} and such that are more \emph{expressive but complex}.
    Because accessibility is a precondition to simulating the parts,
    it is not the same as the transparency level.
    For example, very large, transparent decision trees or very high-dimensional
    (general) linear models may be perceived as globally complex by the end-user.
    However, when looking at the simulatable parts of the explanator,
    like small groups of features or nodes,
    they are easy to grasp.
    \begin{examples}
        Accessibility can indirectly be assessed by the complexity and expressivity
        of the explanation (see \autoref{sec:taxonomy.metrics}).
        To give some examples:
        Simple presentations are, \forexample,
        linear models, general additive models,
        decision trees and Boolean decision rules,
        Bayesian models, or
        clusters of examples (\cf \autoref{sec:taxonomy.problem});
        generally, more complex are, \forexample,
        first-order or fuzzy logical decision rules.
    \end{examples}
    
    \item[Privacy awareness]
    Sensible information like names may be contained in parts of the explanation,
    even though they are not necessary for understanding the actual decision.
    In such cases, an important point is privacy awareness \cite{calegari_integration_2020}:
    Is sensible information removed if unnecessary or properly anonymized if needed?
\end{subparagraphs}


\subsubsection{Interactivity}\label{sec:taxonomy.explanator.interactivity}
The interaction of the user with the explanator may either be static, so the explainee
is once presented with an explanation, or interactive, meaning an iterative process
accepting user feedback as explanation input.
Interactivity is characterized by the \emph{interaction task} and the \emph{explanation process}. 

\begin{subparagraphs}[font=\appenddot]
    \item[Interaction task]
    The user can either inspect explanations or \emph{correct} them.
    Inspecting takes place through \emph{exploration} of different parts of one explanation
    or through consideration of various alternatives and complementing explanations,
    such as implemented in the \emph{iNNvestigate} toolbox \cite{alber_innvestigate_2019}.
    Besides, the user can be empowered within the human-AI partnership to provide corrective feedback to the system via an explanation interface, in order to adapt the explanator and thus the explanandum.
    \begin{examples}
        State-of-the-art systems
        \begin{itemize}
            \item
            enable the user to perform \emph{corrections on labels} and to act upon wrong explanations through interactive machine learning (intML),
            such as implemented in the approach
            \inlineexamplemethod{CAIPI \cite{teso2019explanatory}},
            \item
            they allow for \emph{re-weighting of features} for explanatory debugging, like the system
            \inlineexamplemethod{EluciDebug~\cite{kulesza2010explanatory}},
            \item
            \emph{adaption of features} as provided by
            \inlineexamplemethod{Crayons~\cite{fails2003interactive}},
            and
            \item
            correcting generated verbal explanations through user-defined constraints,
            such as implemented in the medical-decision support system
            \inlineexamplemethod{Learn\-With\-ME~\cite{schmid2020mutual}}.
        \end{itemize}
    \end{examples}
    
    \item[Explanation process]
    As mentioned above, explanation usually takes place in an iterative fashion.
    Sequential analysis allows the user to query further information in an
    iterative manner and to understand the model and its decisions over time,
    in accordance with the users' capabilities and the given context \cite{el2019towards,finzel2021explanation}.
    \begin{examples}
        \examplemethod{Multi-modal explanations~\cite{Hendricks_2018_ECCV}}
        The explanation process includes combining different methods to create multi-modal explanations and involving the user through dialogue. It can be realized in a phrase-critic model as presented by \longcite{Hendricks_2018_ECCV}, or with the help of an explanatory dialogue such as proposed by \longcite{finzel2021explanation}.
    \end{examples}
\end{subparagraphs}


\subsubsection{Mathematical constraints}\label{sec:taxonomy.explanator.constraints}
Mathematical constraints encode some formal properties of the explanator
that were found to be helpful for explanation receival.
Constraints mentioned in the literature are:
\begin{subparagraphs}[font=\appenddot]
    \item[Linearity]
    Considering a concrete proxy model as explanator output,
    linearity is often a desirable form of simplicity
    \cite{kim_interpretability_2018,carvalho_machine_2019,molnar_interpretable_2020}.
    
    \item[Monotonicity]
    Similar to linearity, one here considers a concrete proxy model
    as the output of the explanator.
    The dependency of that model's output on one input feature may be monotonous.
    Monotonicity is desirable for sake of simplicity.
    
    \item[Satisfiability]
    This is the case if the explanator outputs readily allow the application of
    formal methods like solvers.
    
    \item[Number of iterations]
    While some XAI methods require a one-shot inference of the explanandum model
    (\forexample, gradient-based methods), others require several iterations
    of queries to the explanandum.
    Since these might be costly or even restricted in some use cases, a limited
    number of iterations needed by the explanator may be desirable in some cases.
    Such restrictions may arise from non-gameability~\cite{langer_what_2021}
    constraints on the explanandum model, \idest, the number of queries is restricted
    in order to guard against systematic optimization of outputs by users
    (\forexample, searching for adversaries).
    
    \item[Size constraints]
    Many explanator types, respectively surrogate model types, allow for architectural constraints,
    like size or disentanglement, that correlate with reduced complexity.
    See also the respective complexity / interpretability metrics in \autoref{sec:taxonomy.metrics}.
    \begin{examples}
        For linear models, sparsity can reduce complexity \cite{gleicher_framework_2016},
        and for decision trees, depth and width can be constrained.
        One common way to achieve sparsity is to add regularization terms to the training
        of linear explanators and interpretable models.
    \end{examples}
\end{subparagraphs}


\subsection{Metrics}\label{sec:taxonomy.metrics}
\begin{figure}
    \centering
    \includegraphics[scale=\thetaxonomygraphicscale]{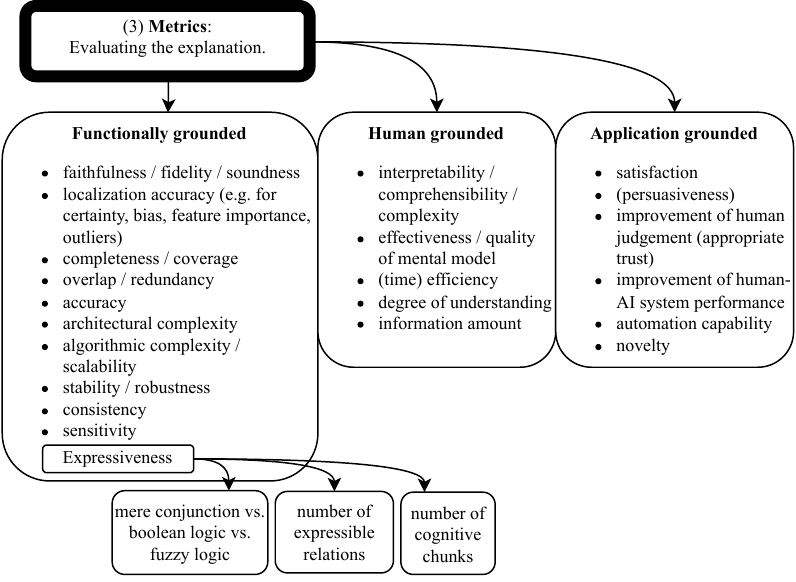}
    \caption{
        Overview of the XAI metrics detailed in \autoref{sec:taxonomy.metrics}.
        Find a visualization of the complete taxonomy in \autoref{fig:taxonomy} on page~\pageref{fig:taxonomy}.
        }
    \label{fig:taxonomy.metrics}
\end{figure}

By now, there is a considerable amount of metrics being suggested to assess the quality
of XAI methods with respect to different goals.
This section details the types of metrics considered in the literature.
Following the original suggestion by \longcite{doshi-velez_rigorous_2017}, we categorize
metrics by their level of human involvement required to measure them:
\begin{itemize}
    \item \emph{\hyperref[sec:taxonomy.metrics.functional]{functionally-grounded}} metrics,
    \item \emph{\hyperref[sec:taxonomy.metrics.human]{human-grounded}} metrics, and
    \item \emph{\hyperref[sec:taxonomy.metrics.application]{application-grounded}} metrics.
\end{itemize}
An overview is given in \autoref{fig:taxonomy.metrics}.
For a selection of approaches to measure the below-described metrics, we refer the reader to 
\longcite{li_quantitative_2020}. They provide a good starting point with an
in-depth analysis of metrics measurement for visual feature attribution methods.

\subsubsection{Functionally-grounded metrics}\label{sec:taxonomy.metrics.functional}
Metrics are considered functionally-grounded if they do not require any human feedback
but instead measure the formal properties of the explanator.
This applies to the following metrics:
\begin{subparagraphs}
    \item[Faithfulness] \cite{li_quantitative_2020}, fidelity~\cite{carvalho_machine_2019}, soundness~\cite{yao_knowledge_2005}, or causality~\cite{calegari_integration_2020},
    measures how accurately the behavior of the explanator conforms with that of the actual object of explanation.
    If a full surrogate model is used, this is the accuracy of the surrogate model outputs with respect to the explanandum outputs.
    And the fidelity of inherently interpretable models also serving as explanators is naturally 100\%.
    Note that fidelity is more often used if the explanator consists of a complete surrogate model (\forexample, a linear proxy like LIME~\cite{ribeiro_why_2016})
    and faithfulness in more general contexts \cite{burkart_survey_2021}.
    Other works use the terms interchangeably like \longcite{du_techniques_2019,carvalho_machine_2019}, which we adopt here.
    More simplification usually comes along with less faithfulness since corner cases
    are not captured anymore, also called the \emph{fidelity-interpretability trade-off}.

    \item[Localization accuracy] with respect to some ground truth
    (\cf \cite{li_quantitative_2020} for visual feature importance maps)
    means how well an explanation correctly localizes
    certain points of interest.
    These points of interest must be given by a ground truth that is preferably
    provided by mathematical properties, such as
    certainty, bias, feature importance, and outliers (\cf \cite{carvalho_machine_2019}).
    In which way such points are highlighted for the explainee depends on the explanation type.
    For example, they could be highlighted in a feature importance heatmap \cite{li_quantitative_2020} or expressed by the aggregated relevance within regions of interest \cite{rieger2020verifying}.
    Note that localization capability is closely related to faithfulness but refers to specific properties of interest.
    
    \item[Completeness,] or coverage,
    measures how large the validity range of an explanation is, so in which subset
    of the input space high fidelity can be expected.
    It can be seen as a generalization of fidelity to the distribution of fidelity.
    Note that coverage can also be calculated for parts of an explanation, such as
    single rules of a rule set, as considered by \longcite{burkart_survey_2021}.

    \item[Overlap]
    is considered by \longcite{burkart_survey_2021} for rule-based explanations.
    It measures the number of data samples that satisfy more than one rule of the rule set,
    \idest, measures the size of areas where the validity ranges of different rules overlap.
    This can be seen as a measure of redundancy in the rule set
    and may sometimes correlate with perceived complexity \cite{burkart_survey_2021}.

    \item[Accuracy] of the surrogate model
    ignores the prediction quality of the original model and only
    measures the prediction quality of the surrogate model for the original task (using the standard accuracy metric).
    This only applies to post-hoc explanations.
    
    \item[Architectural complexity] 
    can be measured using metrics specific to the explanator type.
    The goal is to approximate the subjective, human-grounded complexity that
    a human perceives using purely architectural properties like size measures.
    Such architectural metrics can be, e.g., the number of used input features for feature importance,
    the number of changed features for counterfactual examples (also called \emph{sparsity} \cite{verma_counterfactual_2020}),
    the sparsity of linear models \cite{gleicher_framework_2016},
    the width or depth of decision trees \cite{burkart_survey_2021},
    or, in case of rules, the number of defined rules and the number of unique used predicates \cite{burkart_survey_2021}.

    \item[Algorithmic complexity] and scalability
    measure the information-theoretic complexity of the algorithm used to derive
    the explanator. This includes the time to convergence (to an acceptable solution)
    and is especially interesting for complex approximation schemes like rule extraction.
    
    \item[Stability] or robustness~\cite{calegari_integration_2020}
    measures the change of explanator (output) given a change in the input samples.
    This analogon to (adversarial) robustness of deep neural networks and a stable
    algorithm is usually also better comprehensible and desirable.
    
    \item[Consistency]
    measures the change of the explanator (output) given a change in the model to explain.
    The idea behind consistency is that functionally equivalent models should produce the
    same explanation. This assumption is important for model-agnostic approaches, while
    for model-specific ones, a dependency on the model architecture may even be desirable.
    (\forexample, for architecture visualization).
    
    \item[Sensitivity]
    measures whether local explanations change if the model output changes strongly.
    A big change in the model output usually comes along with a change in the discrimination strategy of the model between the differing samples \cite{li_quantitative_2020}.
    Such changes should be reflected in the explanations.
    Note that this may be in conflict with stability goals for regions in which
    the explanandum model behaves chaotically.

    \item[Expressiveness] or the level of detail
    refers to the level of detail of the formal language used by the explanator.
    It is interested in approximating the expected information density perceived by the user.
    It is closely related to the level of abstraction of the presentation.
    Several functionally-grounded proxies were suggested to obtain comparable
    measures for expressivity:
    \begin{itemize}
        \item the depth or amount of \emph{added information},
        also measured as the mean number of used information units per explanation;
        \item \emph{number of relations} that can be expressed; and
        \item the \emph{expressiveness category} of used rules, namely
        mere conjunction, boolean logic, first-order logic, or fuzzy rules
        (\cf \cite{yao_knowledge_2005}).
    \end{itemize}
\end{subparagraphs}


\subsubsection{Human-grounded metrics}\label{sec:taxonomy.metrics.human}
Other than functionally-grounded metrics, human-grounded metrics require
to involve a human directly on proxy tasks for their measurement.
Human involvement can be measured through observation of a person's reactions but also through direct human feedback.
Often, proxy tasks are considered instead of the final application to avoid
a need for expensive experts or application runtime (think of medical domains).
The goal of an explanation always is that the receiver of the explanation
can build a \emph{mental model} of (aspects of) the object of explanation \cite{kulesza2013too}.
Human-grounded metrics aim to measure some fundamental psychological properties
of the XAI methods, namely quality of the \emph{mental model}.
The following are counted as such in literature:
\begin{subparagraphs}
    \item[Interpretability] or comprehensibility, or complexity
    measures how accurately the mental model approximates the explanator model.
    This measure mostly relies on subjective user feedback on whether they
    \enquote{could make sense} of the presented information.
    It depends on background knowledge, biases, and cognition of the subject
    and can reveal the use of vocabulary inappropriate to the user
    \cite{gilpin_explaining_2018}.
    
    \item[Effectiveness]
    measures how accurately the mental model approximates the object of explanation.
    In other words, one is interested in how well a human can simulate the
    (aspects of interest of the) object after being presented with the explanations.
    Proxies for effectiveness can be fidelity and accessibility
    \cite[Sec.~2.4]{molnar_interpretable_2020}.
    This may serve as a proxy for interpretability.
    
    \item[(Time) efficiency]
    measures how time efficient an explanation is, \idest, how long it takes
    a user to build up a viable mental model.
    This is especially of interest in applications with a limited time frame
    for user reaction, like
    product recommendation systems~\cite{nunes_systematic_2017} or
    automated driving applications~\cite{kim_textual_2018}.
    
    \item[Degree of understanding]
    measures in interactive contexts the current status of understanding.
    It helps to estimate the remaining time or measures needed to reach
    the desired extent of the explainee's mental model.
    
    \item[Information amount]
    measures the total subjective amount of information conveyed by one explanation.
    Even though this may be measured on an information-theoretic basis, it
    usually is subjective and thus requires human feedback.
    Functionally-grounded related metrics are the (architectural) complexity of the object of explanation, together with fidelity and coverage.
    For example, more complex models have a tendency to contain more information,
    and thus require more complex explanations if they are to be approximated
    widely and accurately.
\end{subparagraphs}


\subsubsection{Application-grounded metrics}\label{sec:taxonomy.metrics.application}
Other than human-grounded metrics, application-grounded ones work on human feedback
for the final application.
The following metrics are considered application-grounded:
\begin{description}
    \item[Satisfaction]
    measures the direct content of the explainee with the system. It implicitly measures the benefit of explanations for the explanation system user.
    
    \item[Persuasiveness]
    assesses the capability of the explanations to nudge an explainee into a certain
    direction.
    This is foremostly considered in recommendation systems \cite{nunes_systematic_2017}
    but has high importance when it comes to analysis tasks, where false positives and
    false negatives of the human-AI system are undesirable.
    In this context, a high persuasiveness may indicate a miscalibration of indicativeness.
    
    \item[Improvement of human judgment] \cite{mueller_principles_2021}
    measures whether the explanation system user develops an appropriate level of
    trust in the decisions of the explained model.
    Correct decisions should be trusted more than wrong decisions,
     \forexample because explanations of wrong decisions are illogical.
    \item[Improvement of human-AI system performance]
    considers the end-to-end task to be achieved by all of the following: explanandum, explainee, and explanator.
    This can, \forexample, be the diagnosis quality of doctors assisted by a recommendation system
    \cite{mueller_principles_2021,schmid2020mutual}.
    
    \item[Automation capability]
    gives an estimate of how much of the manual work conducted by the human
    in the human-AI system can be automatized.
    Especially for local explanation techniques, automation may be an important
    factor for feasibility if the number of samples a human needs to scan can
    be drastically reduced \cite{finzel2021deriving}.
    
    \item[Novelty]
    estimates the subjective degree of novelty of information provided to the
    explainee \cite{langer_what_2021}.
    This is closely related to efficiency and satisfaction:
    Especially in exploratory use cases, high novelty can drastically
    increase efficiency (no repetitive work for the explainee) and
    keep satisfaction high (decrease the possibility of boredom for the explainee).
\end{description}

\begingroup
\footnotesize
\newcolumntype{f}{@{~}c}
\newcommand{\cluster}[1]{\\\secseprule{9}\multicolumn{9}{@{}l@{}}{\textbf{#1}}}
\let\cite\longcite
\begin{tabularx}{\linewidth}{@{} >{\raggedright}p{8em} >{\raggedright}X @{} c@{}cfffff@{}}
    \caption{Review of an exemplary selection of XAI techniques according
    to the defined taxonomy aspects (without inherently transparent models from \autoref{sec:taxonomy.problem.interpretability}).
    Abbreviations by column:
    \abbr{image data}{img}, \abbr{point cloud data}{pcl};
    \abbr{Trans.}{transparency}, \abbr{post-hoc}{p}, \abbr{transparent}{t}, \abbr{self-explaining}{s}, \abbr{blended}{b};
    \abbr{processing}{p}, \abbr{representation}{r}, \abbr{development during training}{t} \abbr{data}{d};
    \abbr{visual}{vis}, \abbr{symbolic}{sym}, \abbr{plot}{plt};
    \abbr{feature importance}{fi}, \abbr{contrastive}{con}, \abbr{prototypical}{proto}, \abbr{decision tree}{tree}, \abbr{distribution}{dist}}
    \label{tab:methods.overview}\\ 
    	\rot{Name}	&	\rot{Cite}	&	\rot{Task}	&	\rot{Model-agnostic?}	&	\rot{Transp.}	&	\rot{Global?}	&	\rot{Obj. Expl.}	&	\rot{Form}	&	\rot{Type}	%
\endfirsthead\caption{continued from page~\pageref{tab:methods.overview}}\\																		\rot{Name}	&	\rot{Cite}	&	\rot{Task}	&	\rot{Model-agnostic?}	&	\rot{Transp.}	&	\rot{Global?}	&	\rot{Obj. Expl.}	&	\rot{Form}	&	\rot{Type}	%
\endhead	
\cluster{Self-explaining and blended models}%
\\ \midrule	-	&	\cite{hendricks_generating_2016}	&	cls	&		&	s	&		&	p	&	sym/vis	&	rules/fi	%
\\ \midrule	-	&	\cite{kim_textual_2018}	&	any	&		&	s	&		&	p	&	sym/vis	&	rules/fi	%
\\ \midrule	ProtoPNet	&	\cite{chen_this_2019}	&	cls,img	&		&	s	&		&	p/r	&	vis	&	proto/fi	%
\\ \midrule	Capsule Nets	&	\cite{sabour_dynamic_2017}	&	cls	&		&	s	&		&	r	&	sym	&	fi	%
\\ \midrule	Semantic Bottlenecks, ReNN, Concept Whitening	&	\cite{losch_interpretability_2019,wang_renn_2018,chen_concept_2020}	&	any	&		&	s	&		&	r	&	sym	&	fi	%
\\ \midrule	Logic Tensor Nets	&	\cite{donadello_logic_2017}	&	any	&		&	b	&	\checkmark	&	p/r	&	sym	&	rule	%
\\ \midrule	FoldingNet	&	\cite{yang_foldingnet_2017}	&	any,pcl	&		&	b	&		&	p	&	vis	&	fi/red	%
\\ \midrule	Neuralized clustering	&	\cite{kauffmann_clustering_2019}	&	any	&		&	b	&		&	p	&	vis	&	fi	%
\cluster{Black-box heatmapping}
\\ \midrule	LIME, SHAP	&	\cite{ribeiro_why_2016,lundberg_unified_2017}	&	cls	&	\checkmark	&	p	&		&	p	&	vis	&	fi/con	%
\\ \midrule	RISE	&	\cite{petsiuk_rise_2018}	&	cls,img	&	\checkmark	&	p	&		&	p	&	vis	&	fi	%
\\ \midrule	D-RISE	&	\cite{petsiuk_blackbox_2021}	&	det,img	&	\checkmark	&	p	&		&	p	&	vis	&	fi	%
\\ \midrule	CEM	&	\cite{dhurandhar_explanations_2018}	&	cls,img	&	\checkmark	&	p	&		&	p	&	vis	&	fi/con	%
\cluster{White-box heatmapping}
\\ \midrule	Sensitivity analysis	&	\cite{baehrens_how_2010}	&	cls	&		&	p	&		&	p	&	vis	&	fi	%
\\ \midrule	Deconvnet, (Guided) Backprop.	&	\cite{zeiler_visualizing_2014,simonyan_deep_2014,springenberg_striving_2015}	&	img	&		&	p	&		&	p	&	vis	&	fi	%
\\ \midrule	CAM, Grad-CAM	&	\cite{zhou_learning_2016,selvaraju_gradcam_2017}	&	cls,img	&		&	p	&		&	p	&	vis	&	fi	%
\\ \midrule	SIDU	&	\cite{muddamsetty_introducing_2021}	&	cls,img	&		&	p	&		&	p	&	vis	&	fi	%
\\ \midrule	Concept-wise Grad-CAM	&	\cite{zhou_interpretable_2018}	&	cls,img	&		&	p	&		&	p/r	&	vis	&	fi	%
\\ \midrule	SIDU	&	\cite{muddamsetty_introducing_2021}	&	cls,img	&		&	p	&		&	p	&	vis	&	fi	%
\\ \midrule	LRP	&	\cite{bach_pixelwise_2015}	&	cls	&		&	p	&		&	p	&	vis	&	fi	%
\\ \midrule	Pattern Attribution	&	\cite{kindermans_learning_2018}	&	cls	&		&	p	&		&	p	&	vis	&	fi	%
\\ \midrule	-	&	\cite{fong_interpretable_2017}	&	cls	&		&	p	&		&	p	&	vis	&	fi	%
\\ \midrule	SmoothGrad, Integrated Gradients	&	\cite{smilkov_smoothgrad_2017,sundararajan_axiomatic_2017}	&	cls	&		&	p	&		&	p	&	vis	&	fi	%
\\ \midrule	Integrated Hessians	&	\cite{janizek_explaining_2020}	&	cls	&		&	p	&		&	p	&	vis	&	fi	%
\cluster{Global representation analysis}
\\ \midrule	Feature Visualization	&	\cite{olah_feature_2017}	&	img	&		&	p	&	\checkmark	&	r	&	vis	&	proto	%
\\ \midrule	NetDissect	&	\cite{bau_network_2017}	&	img	&		&	p	&	\checkmark	&	r	&	vis	&	proto/fi	%
\\ \midrule	Net2Vec	&	\cite{fong_net2vec_2018}	&	img	&		&	p	&	(\checkmark)	&	r	&	vis	&	fi	%
\\ \midrule	TCAV	&	\cite{kim_interpretability_2018}	&	any	&		&	p	&	\checkmark	&	r	&	vis	&	fi	%
\\ \midrule	ACE	&	\cite{ghorbani_automatic_2019}	&	any	&		&	p	&	\checkmark	&	r	&	vis	&	fi	%
\\ \midrule	-	&	\cite{yeh_completenessaware_2020}	&	any	&		&	p	&	\checkmark	&	r	&	vis	&	proto	%
\\ \midrule	IIN	&	\cite{esser_disentangling_2020}	&	any	&		&	p	&	(\checkmark)	&	r	&	vis/sym	&	fi	%
\\ \midrule	Explanatory Graph	&	\cite{zhang_interpreting_2018}	&	img	&		&	p	&	(\checkmark)	&	p/r	&	vis	&	graph	%
\cluster{Dependency plots}
\\ \midrule	PDP	&	\cite{friedman_greedy_2001}	&	any	&	\checkmark	&	p	&		&	p	&	vis	&	plt	%
\\ \midrule	ICE	&	\cite{goldstein_peeking_2015}	&	any	&	\checkmark	&	p	&	\checkmark	&	p	&	vis	&	plt	%
\cluster{Rule extraction}
\\ \midrule	TREPAN, C4.5, Concept Tree	&	\cite{craven_extracting_1995,quinlan_c4_1993,renard_concept_2019}	&	cls	&	\checkmark	&	p	&	\checkmark	&	p	&	sym	&	tree	%
\\ \midrule	VIA	&	\cite{thrun_extracting_1995}	&	cls	&	\checkmark	&	p	&	\checkmark	&	p	&	sym	&	rules	%
\\ \midrule	DeepRED	&	\cite{zilke_deepred_2016}	&	cls	&		&	p	&	\checkmark	&	p	&	sym	&	rules	%
\\ \midrule	LIME-Aleph	&	\cite{rabold_explaining_2018}	&	cls	&	\checkmark	&	p	&		&	p	&	sym	&	rules	%
\\ \midrule	CA-ILP	&	\cite{rabold_expressive_2020}	&	cls	&		&	p	&	\checkmark	&	p	&	sym	&	rules	%
\\ \midrule	NBDT	&	\cite{wan_nbdt_2020}	&	cls	&		&	p	&	\checkmark	&	p	&	sym	&	tree	%
\cluster{Interactivity}
\\ \midrule	CAIPI	&	\cite{teso2019explanatory}	&	cls,img	&	\checkmark	&	p	&		&	r	&	vis	&	fi/con	%
\\ \midrule	EluciDebug	&	\cite{kulesza2010explanatory}	&	cls	&	\checkmark	&	p	&		&	r	&	vis	&	fi,plt	%
\\ \midrule	Crayons	&	\cite{fails2003interactive}	&	cls,img	&	\checkmark	&	t	&		&	p	&	vis	&	plt	%
\\ \midrule	LearnWithME	&	\cite{schmid2020mutual}	&	cls	&	\checkmark	&	t	&	\checkmark	&	p, r	&	sym	&	rules	%
\\ \midrule	Multi-modal phrase-critic model	&	\cite{Hendricks_2018_ECCV}	&	cls,img	&		&	p	&	\checkmark	&	p	&	vis,sym	&	plt,rules	%
\cluster{Inspection of the training}
\\ \midrule	-	&	\cite{shwartz-ziv_opening_2017}	&	any	&		&	p	&	\checkmark	&	t	&	vis	&	dist	%
\\ \midrule	Influence functions	&	\cite{koh_understanding_2017}	&	cls	&		&	p	&	\checkmark	&	t	&	vis	&	fi/dist	%
\cluster{Data analysis methods}
\\ \midrule	t-SNE, PCA	&	\cite{maaten_visualizing_2008,jolliffe_principal_2002}	&	any	&	\checkmark	&	p	&	\checkmark	&	d	&	vis	&	red	%
\\ \midrule	k-means, spectral clustering	&	\cite{hartigan_algorithm_1979,vonluxburg_tutorial_2007}	&	any	&	\checkmark	&	p	&	\checkmark	&	d	&	vis	&	proto	%
\\ \bottomrule																		
\end{tabularx}
\normalsize
\endgroup


\section{Discussion and conclusion}
\label{sec:conclusion}
The abundance of existing literature on structuring the field and methods of XAI has by now reached an overwhelming level for beginners and practitioners.
To help in finding good starting points for a deeper dive, we here presented a rich and structured survey of surveys on XAI topics.
This showed an increasing breadth of application fields and method types investigated in order to provide explainable yet accurate learned models.
Especially, the exponentially increasing number of both XAI methods and method surveys suggests an increasing interest in XAI unrestrained growth of the research field in the upcoming years.
As some persisting trends, we found the application domains of medicine and recommendation systems, as well as (visual) explanations for visual models.
Some upcoming or re-awakening trends seem to be the field of natural language processing, and rule-based explanation methods.

With the ever-growing amount of methods for achieving explainability and interpretability, XAI method surveys developed many approaches for method categorization.
Early foundational concepts have been extended and sub-structured to increasingly detailed collections of aspects for differentiating XAI methods.
This paper systematically united aspects scattered over existing literature into an overarching taxonomy structure.
Starting from the definition of the problem of XAI, we found the following three main parts of an explanation procedure suitable for a top-level taxonomy structure: the task, the explainer, and evaluation metrics.
This paper in detail defines and sub-structures each of these parts.
The applicability of those taxonomy aspects for differentiating methods is evidenced practically:
Numerous example methods from the most relevant as well as the most recent research literature are discussed and categorized according to the taxonomy aspects.
An overview of the examples is given in the end, concentrating on the seven classification criteria that are most significant in the literature. These concretely are%
the \emph{task},
the form of \emph{interpretability} (\forexample, inherently interpretable),
whether the method is \emph{model-agnostic or model-specific},
whether it generates \emph{global or local} explanations,
what the \emph{object of explanation} is,
in what form explanations are \emph{presented}, and
the type of explanation.

As highlighted in the course of the paper, the creation of an explanation system
should be tailored tightly to the use-case: This holds for all of development, application, and evaluation of XAI the system.
Concretely, the different stakeholders and their contexts should be taken into account \cite{langer_what_2021,gleicher_framework_2016}.
Our proposed taxonomy may serve as a profound basis to analyze stakeholder needs and formulate concrete, use-case-specific requirements upon the explanation system.
The provided survey of surveys then provides the entry point when looking for XAI methods fulfilling the derived requirements.

Altogether, our proposed unified taxonomy and our survey allow
(a)~beginners to gain an easy and targeted overview and entry point to the field of XAI;
(b)~practitioners to formulate sufficient requirements for their explanation use-case and to find accordingly suitable XAI methods; and lastly
(c)~researchers to properly position their research efforts on XAI methods and identify potential gaps.
We hope this work fosters research and fruitful application of XAI.

Finally, it must be mentioned that our survey is and will not be or stay complete: Despite our aim for a broad representation of the XAI field, the used structured literature search is naturally biased by the current interests in specific sub-fields of AI and the search terms. On the other hand, we hope this ensures relevance to a large part of the research community.
Furthermore, we concentrated on a broadly applicable taxonomy for XAI methods, whilst sub-fields of XAI may  prefer or come up with more detailed differentiation aspects or a different taxonomy structure.

We are looking forward to seeing future updates of the state of research captured in this work and practical guides for choosing the right XAI method according to a use-case definition.

\section*{Declarations}
\begin{description}[font=\normalfont\bfseries\appenddot, leftmargin=0pt, parsep=.2\baselineskip]
\item[Funding and employment]
Funding for this research was obtained as declared in the acknowledgments, with the first author employed at the Continental Automotive GmbH, Germany,
and the second at the University of Bamberg, Germany.

\item[Competing interests]
This work was authored in the scope of doctoral research of both authors, both supervised by Prof.\,Dr.\,Ute Schmid at the University of Bamberg.
The authors declare that they have no competing financial interests.

\item[Ethics approval and consent to participate] Not applicable.
\item[Data/material/code availability] Not applicable.

\item[Authors’ contribution statement] No contributors were involved other than the declared authors.
\end{description}


\bibliography{literature}

\end{document}